\documentclass{article}
\usepackage{graphicx} 
\usepackage[table]{xcolor}
\usepackage{tikz}
\usepackage{amsfonts}
\usepackage{amssymb}
\usepackage{amsmath}
\usepackage{multirow}
\usepackage{array}
\usepackage{booktabs}
\usepackage{float}
\usepackage{longtable}
\usepackage{subcaption}
\usepackage{arydshln}
\usepackage{rotating}
\usepackage{appendix}

\definecolor{cs4}{RGB}{215,48,39}
\definecolor{cs5}{RGB}{244,109,67}
\definecolor{cs6}{RGB}{253,174,97}
\definecolor{cs7}{RGB}{254,224,144}
\definecolor{cs8}{RGB}{217,239,139}
\definecolor{cs9}{RGB}{145,207,96}
\definecolor{cs10}{RGB}{26,152,80}
\definecolor{mG}{HTML}{4C72B0}
\definecolor{mP}{HTML}{DD8452}
\definecolor{mD}{HTML}{55A868}
\definecolor{mS}{HTML}{C44E52}
\definecolor{mI}{HTML}{8172B2}
\definecolor{mF}{HTML}{937860}
\definecolor{catpre}{RGB}{228,240,255}
\definecolor{cattra}{RGB}{220,248,232}
\definecolor{catgen}{RGB}{255,242,218}
\definecolor{catllm}{RGB}{240,228,255}
\definecolor{catnlp}{RGB}{255,226,226}
\usepackage{biblatex} 
\usepackage[a4paper, left=20mm, right=20mm, top=20mm, bottom=25mm]{geometry}

\newcommand{\system}{\mbox{OpTFM}}

\title{A Comparative Framework for Evaluating Foundation Models on Tabular Data: A Case Study in Healthcare}
\author{Majid Lotfian Delouee, Sjors G. J. G. In ’t Veld, Martijn C. Schut}

\date{}
\begin{document}

\maketitle
\begin{abstract}
Tabular data is the most common format in clinical practice, encompassing laboratory results, medication records, diagnostic codes, and patient demographics. As foundation models for tabular data have grown in number and variety, a practical question has become harder to answer: which model should a clinician or data scientist actually choose for a given task, and why? Existing surveys catalogue what these models can do, but they stop short of providing a structured way to compare them against the specific demands of a real application. We introduce \system{}, a comparative evaluation framework that scores and ranks tabular foundation models (TFMs) across six clinically meaningful dimensions: how well a model generalizes to new datasets, how effectively it protects patient privacy, how much data it needs to perform well, how it scales with growing datasets and feature spaces, how interpretable its predictions are to clinicians, and how fairly it performs across patient subgroups. Each dimension is broken down into measurable sub-components, and groups of sub-components can optionally be combined into supplementary compound scores, called super-metrics, that provide a diagnostic view of how a model performs across several dimensions simultaneously. To show how the framework works in practice, we apply it to two healthcare use cases, screening for iron deficiency and predicting heart failure, demonstrating how the same set of metrics leads to different model rankings depending on what matters most in each clinical context. We also provide a taxonomy of 45 TFMs organized by their underlying architecture, which serves as a reference for researchers and practitioners looking to navigate this rapidly expanding field.
\end{abstract}

\section{Introduction}
\label{sec:introduction}

Clinical databases are built on tabular data. Every time a patient has a blood test, receives a diagnosis, or starts a new medication, that information is recorded in rows and columns, whether in an electronic health record, a hospital information system, or a research registry. This format is not incidental; it reflects how medicine organizes knowledge. Features like age, hemoglobin level, ejection fraction, or creatinine carry meaning precisely because they can be compared across patients, tracked over time, and correlated with outcomes. The structured nature of tabular data is what makes it analyzable at scale, and it is what allows researchers to ask questions like which laboratory values best predict readmission, or which patient profiles respond best to a given treatment.

Natural language processing and deep learning have transformed the analysis of unstructured clinical data, such as free-text notes and radiology reports, and those advances have carried over into how we handle structured data as well. Recurrent neural networks and transformer-based architectures, originally developed for language, have been adapted to learn from sequences of clinical events, mixed numerical and categorical features, and heterogeneous data types that traditional machine learning methods struggle to handle in an integrated way. In healthcare settings, these models have been applied to named entity recognition in clinical text, sentiment analysis of patient feedback, and automated question answering over medical records, among other tasks.

The emergence of large language models has added another layer to this landscape. These models, trained on vast collections of text and code, can generate coherent clinical summaries, assist with differential diagnosis, and support structured data analysis in ways that were not practical even a few years ago. Beyond their text-based applications, large language models are beginning to influence how we approach tabular data more directly, partly because they can generate realistic synthetic records to address data scarcity, and partly because their pre-training provides a rich prior that can improve performance on downstream tasks when labeled data is limited.

Foundation models represent the broader class of systems built on this paradigm. Unlike earlier task-specific models, a foundation model is trained on large and diverse data to develop general-purpose representations that can be adapted to many different downstream applications. In healthcare, this means a single pre-trained model could potentially support clinical risk prediction, patient stratification, treatment response modeling, and operational analytics, all from the same underlying architecture, with domain-specific fine-tuning applied as needed. The appeal is significant: training a specialized model from scratch for each clinical task is costly, time-consuming, and often constrained by the limited availability of labeled data in medical settings.

Research on foundation models for tabular data has grown considerably in recent years. Several studies have surveyed the landscape, categorizing models by the structure of the data they handle, the tasks they are designed for, and the learning strategies they employ \cite{ruan2024language,zhang2023towards,fang2024large,breugel2024position}. These surveys have been valuable for understanding the evolution of the field and for identifying the key technical challenges involved in applying language model ideas to structured data. They make clear, for instance, that tabular data presents unique difficulties, including mixed feature types, missing values, and the lack of natural sequential ordering, that require purpose-built solutions rather than direct transfer from text or image domains.

What these surveys do not provide, however, is a practical tool for model selection. They describe what exists, but they do not help a clinician or data scientist decide which model to use for a specific task. The gap is not trivial. Some models prioritize privacy by design, which matters enormously in healthcare. Others are optimized for speed or for handling high-dimensional feature spaces, which matters more in industrial or financial applications. Some generalize well across datasets with different schemas, while others depend on consistent feature definitions. Without a way to systematically compare models on the dimensions that actually matter for a given use case, practitioners are left to make selection decisions based on incomplete information, benchmark results from tasks that may not reflect their own, or simply familiarity with a particular model.

A further complication is that real deployment decisions rarely ask about a single dimension. A hospital deploying a model across multiple sites does not ask only whether it generalizes; it asks whether it generalizes while tolerating schema differences and resisting patient re-identification attacks at the same time, because all three conditions will arise simultaneously. A clinical procurement team does not evaluate privacy and computational efficiency as independent concerns; it needs to know whether a model can be fast enough for real-time clinical decision support while still meeting GDPR or HIPAA requirements, because a model that excels on one but fails on the other is not deployable. These compound questions require an evaluation framework that can aggregate across dimensions, not just report them separately.

A second gap concerns the categorization of models by architecture. The distinction between encoder-only, decoder-only, and encoder-decoder architectures is not merely technical; it determines what kinds of tasks a model is suited for, how it processes inputs, and what trade-offs it makes between generalization and generation. Existing surveys tend to treat all foundation models as a single category, which obscures these differences and makes it harder to reason about which architectural approach is appropriate for a given clinical application.

This work addresses both gaps. We propose \system{}, a comparative framework that allows researchers and practitioners to evaluate and rank tabular foundation models according to a structured set of criteria, adapted to the priorities of their specific application domain. The framework is built around a three-layer architecture. At the base, sub-metrics are operationally defined, measurable properties, each anchored to a specific empirical test. One layer up, six metrics aggregate sub-metrics into domain-organized profiles covering generalizability, privacy, data efficiency, scalability, interpretability, and fairness. An optional super-metric layer combines sub-metrics from several metric domains into supplementary compound diagnostic scores that provide additional analytical depth alongside the primary ranking. To ground the framework in practice, we apply it to healthcare and work through two clinical use cases in detail. We also provide a taxonomy of current tabular foundation models organized by architecture.

The specific contributions of \system{} are as follows:
 \begin{enumerate}
    \item A taxonomy of representative tabular foundation models organized by architectural design, distinguishing encoder-only, decoder-only, and encoder-decoder approaches and their practical implications for clinical and applied AI use cases.
    \item A three-layer evaluation architecture in which sub-metrics provide measurable atomic scores, metrics aggregate them into six domain-organized performance dimensions with adjustable weights, and an optional super-metric layer combines sub-metrics across domains into four supplementary compound diagnostic scores (Transfer Learning Capacity, Model Resilience, Efficiency-Privacy Trade-off, and Clinical Trustworthiness) that provide additional analytical depth without replacing the overall ranking as the primary recommendation.
    \item A detailed application of the framework to the healthcare domain, demonstrated through two clinical case studies, iron deficiency screening and heart failure prediction, showing how domain-specific priorities translate into differentiated model rankings and actionable deployment recommendations.
 \end{enumerate}


\section{Background and Related Work}
\label{sec:relatedwork}

The term \textit{foundation model} was introduced by Bommasani and colleagues \cite{bommasani2021opportunities} to describe a new class of AI systems trained on large and diverse datasets, capable of being adapted to a wide range of downstream tasks with relatively little additional training. The defining characteristic of these models is not their size alone, but the breadth of the representations they develop during pre-training. By learning from vast amounts of data before being applied to any specific task, they build a general understanding of patterns, relationships, and structure that can transfer across settings. Early and influential examples include BERT \cite{kenton2019bert}, which learns bidirectional representations of text, and GPT \cite{radford2018gpt}, which learns to generate text autoregressively. Models like GPT-3 \cite{brown2020language} extended this further, demonstrating that scale alone could unlock capabilities in reasoning, translation, and question answering that had not been anticipated. CLIP \cite{radford2021learning} showed that the same paradigm could bridge image and text understanding. In healthcare and the life sciences, this foundation model approach has been applied to tasks ranging from medical image interpretation and drug discovery to clinical text summarization and genomic sequence modeling.

\subsection{Fundamentals of Foundation Models}

The architecture that makes modern foundation models possible is the transformer \cite{vaswani2017attention}. Transformers rely on attention mechanisms that allow a model to weigh the relevance of different parts of its input when making predictions, rather than processing the input sequentially from start to finish. This makes them particularly effective at capturing long-range dependencies in data, a critical property for language, where meaning often depends on context spread across many tokens, and for structured data, where relationships between features can be non-local and non-obvious. The pre-training phase is central to how foundation models acquire their general knowledge. During pre-training, the model is exposed to large, diverse datasets and trained to perform a self-supervised task, predicting masked tokens, reconstructing corrupted inputs, or modeling the distribution of next tokens, without requiring human-labeled examples. The representations learned during this phase form the foundation that downstream fine-tuning builds on. Fine-tuning involves training the pre-trained model further on a smaller, task-specific dataset, adjusting its weights so that the general knowledge it has acquired is aligned with the specific patterns of the new task. In medical settings, this typically means taking a model pre-trained on broad biomedical data and fine-tuning it on a dataset from a specific hospital, disease area, or clinical workflow. The efficiency of this process, relative to training a specialized model from scratch, is one of the central practical arguments for the foundation model approach in data-scarce domains like medicine.

\subsection{Foundation Models for Tabular Data}

Tabular data dominates healthcare informatics. Patient records, laboratory systems, pharmacy databases, and administrative systems all store their core information in row-and-column format, where each row represents a patient or an encounter and each column represents a clinical variable. Despite this prevalence, the application of foundation models to tabular data has lagged behind their application to text and images, partly because the challenges are different in kind. Text has a natural sequential structure that transformers were designed to exploit. Images have spatial structure that convolutional and attention-based architectures can learn to navigate. Tabular data has neither. Columns may be numerical or categorical, present or missing, semantically rich or operationally redundant. There is no inherent ordering of features that the model can leverage, and the meaning of a value depends heavily on which column it comes from.

A growing body of work has adapted the foundation model paradigm specifically for these challenges. Models including TransTab \cite{wang2022transtab}, TabNet \cite{arik2021tabnet}, TabTransformer \cite{huang2020tabtransformer}, NODE \cite{Popov2020Neural}, FT-Transformer \cite{gorishniy2021revisiting}, SAINT \cite{somepalli2022saint}, and TabPFN \cite{hollmann2023tabpfn} each represent a different approach to this problem. More recent work has continued to push the frontier: TabPFN v2 \cite{hollmann2025tabpfnv2} was shown in a Nature study to outperform all previous methods on datasets with up to 10,000 samples using substantially less training time; CARTE \cite{kim2024carte} proposed a graph-attentional architecture that leverages open-vocabulary string embeddings of column names and values to enable cross-table transfer; TabDPT \cite{ma2024tabdpt} combined in-context-learning-based retrieval with self-supervised pre-training on real tabular data; XTab \cite{zhu2023xtab} demonstrated cross-table pre-training of FT-Transformer using federated learning across 84 heterogeneous datasets; TabICL \cite{qu2025tabicl} introduced a two-stage column-then-row attention mechanism that achieves parity with TabPFN v2 while handling datasets with up to 500,000 samples; and UniTabE \cite{yang2024unitabe} proposed a unified tokenizer that maps heterogeneous tabular data across tables into a shared embedding space. TabRet \cite{onishi2023tabret} addresses the specific challenge of unseen columns at inference time through a retokenizing step before fine-tuning. Most recently, TabPFN 3 \cite{hollmann2026tabpfn3} extended the prior-fitted network paradigm to datasets with up to one million training rows and introduced test-time compute scaling to tabular foundation models. TransTab, for example, encodes both column names and cell values as text, allowing the model to transfer across datasets with different schemas by relying on the semantic content of feature names. TabNet introduces a sequential attention mechanism that dynamically selects which features to focus on at each step, making it interpretable in the sense that feature selection decisions can be visualized. SAINT combines row-level and column-level attention, allowing it to capture interactions both within a row and across similar rows in the dataset. TabPFN takes a different direction entirely, framing classification as in-context learning and performing inference without any gradient-based fine-tuning.

The core advantage of these models over classical machine learning approaches is their ability to learn complex feature interactions without manual feature engineering, and their potential to transfer knowledge across datasets. A model pre-trained on diverse tabular data from multiple sources may arrive at a new clinical dataset with better initial representations than one trained from scratch, particularly when labeled data for the target task is limited, a situation that is common in medicine. That said, the limitations are real. Pre-training tabular models requires large collections of high-quality structured data, which are harder to assemble than text corpora and are often fragmented across institutions. The lack of a natural self-supervised pre-training objective, analogous to masked language modeling in NLP, means that different TFMs make quite different assumptions about how to pre-train, and it is not clear which approach generalizes best across clinical domains.

\subsection{Input and Output Representations}

A fundamental design choice for any tabular foundation model is how it represents its inputs and outputs. On the input side, models must handle two qualitatively different types of features: numerical variables, such as a patient's age, serum ferritin level, or left ventricular ejection fraction, and categorical variables, such as diagnosis codes, medication names, or binary clinical flags. Simpler approaches tokenize categorical values through one-hot encoding or lookup tables before passing them to the model. More recent approaches learn dedicated embedding layers that map each category, and sometimes each numerical value through binning or projection, into a shared continuous representation space. Some models go further by treating column names as part of the input, so that a feature called ``serum creatinine'' and one called ``creatinine level'' are represented as semantically related, enabling better generalization when the same clinical concept appears under different names across datasets.

On the output side, TFMs target a range of clinical tasks. Classification heads produce probability estimates for discrete outcomes, whether binary, such as disease-present or disease-absent, or multi-class, such as disease severity staging. Regression heads produce continuous scores, for example a predicted lab value or a risk index. Some models produce row-level embeddings as their primary output, which can then feed into task-specific downstream components. Generative models, particularly those based on decoder architectures, produce entire rows or tables as outputs, which is useful for creating synthetic patient records or for augmenting small training datasets in privacy-constrained settings.

\subsection{Architectural Taxonomy}

Most tabular foundation models are built on the transformer, but the way the transformer is configured determines a great deal about what the model can and cannot do well.

\textbf{Encoder-only models} read the full input at once and build a bidirectional representation of it, meaning each feature's representation can be influenced by all other features simultaneously. This design is well matched to discriminative tasks like classification and regression, where the goal is to produce an informative summary of a given patient record rather than to generate new data. Models such as TabTransformer and BERT-based tabular encoders including TaBERT and TABBIE take this approach. Encoder-only models are computationally efficient at inference time and straightforward to fine-tune, making them practical for clinical deployment. Their limitation is that they cannot generate new data and are not suited to tasks that require producing outputs in sequence.

\textbf{Decoder-only models} generate their outputs one step at a time, with each step conditioned on everything that came before. In the tabular setting, this typically means the model serializes a row as a text sequence and generates new rows by sampling from that distribution. GReaT and TabuLa are examples of this design. Decoder-only models are particularly useful for synthetic data generation, which matters in healthcare because real patient data cannot always be shared across institutions or used freely for model training. They can also perform zero-shot inference on new feature combinations without any fine-tuning. The cost of this flexibility is computational: autoregressive generation is slow, which can be a barrier in latency-sensitive clinical applications.

\textbf{Encoder-decoder models} combine a bidirectional encoder with a step-by-step decoder, allowing the model to understand structured inputs and generate structured outputs. This architecture is well suited to transformation tasks, converting a table to a text summary, parsing a natural language query into a structured answer over a table, or translating between different schema representations. TAPEX and TUTA use variants of this design. These models tend to be larger and more complex than encoder-only approaches, but they support a richer range of tasks relevant to clinical knowledge extraction.

Some influential tabular models do not use the transformer at all. TabNet performs feature selection through a sequential attention mechanism, while NODE constructs differentiable oblivious decision trees. These models often perform competitively with transformer-based approaches on standard benchmarks, particularly in low-data settings, but they do not support the same pre-training and transfer learning workflows that make transformer-based foundation models attractive in data-scarce clinical environments.

\subsection{Models Evaluated in This Work}

The framework is applicable to any tabular foundation model, and Appendix~\ref{sec:appendix} provides an extended taxonomy covering 45 models across all architectural categories. For the healthcare case studies in Section~\ref{sec:healthcare}, eight models were selected as representative examples that collectively span the main design dimensions of the field. FT-Transformer \cite{gorishniy2021revisiting} is a clean encoder-only transformer baseline that has been consistently competitive across benchmarks, requiring no architectural specialization. TabNet \cite{arik2021tabnet} was included because its sequential attention mechanism provides a built-in form of interpretability through feature selection masks, which is relevant in clinical contexts where feature importance is expected by practitioners. TabTransformer \cite{huang2020tabtransformer} uses column-level embeddings that are initialized from categorical inputs and refined through self-attention, representing the class of models that process tabular features with minimal reliance on textual encoding. TransTab \cite{wang2022transtab} was chosen as the representative cross-table model because it encodes both column names and cell values as text, making it capable of transferring across datasets with different schemas without feature alignment, a capability directly relevant to multi-site clinical deployment. MediTab \cite{he2024foundation} is the only model in the comparison specifically designed for healthcare tabular data, with a pre-training pipeline that incorporates clinical reasoning and a data collection strategy oriented toward medical datasets; it serves as the example of a domain-adapted foundation model. UniTabE \cite{yang2024unitabe} was included as the representative of unified tokenizer approaches, mapping heterogeneous tabular features across tables into a shared embedding space. TabPFN~3 \cite{hollmann2026tabpfn3} and TabICL \cite{qu2025tabicl} were included to represent the in-context learning paradigm: TabPFN~3 extends prior-fitted networks to datasets with up to one million training rows, while TabICL introduces a two-stage column-then-row attention mechanism that scales to 500,000 samples. Together, these eight models cover encoder-only and hybrid architectures, span the range from task-agnostic to domain-specific design, and reflect meaningfully different assumptions about privacy, interpretability, scalability, and schema flexibility that the framework is designed to discriminate between.

\subsection{Downstream Tasks}

Tabular foundation models are applied to a broad range of downstream tasks, and understanding which task a model is designed for is an important part of selecting the right model for a clinical application. Supervised classification and regression remain the most common use cases, covering predictions like in-hospital mortality, disease diagnosis, readmission risk, and treatment response. Few-shot and zero-shot inference are growing in importance, particularly for smaller institutions that do not have the labeled data needed to fine-tune a full model. Synthetic data generation allows institutions to create realistic patient records that preserve statistical properties without exposing individual identities, which is particularly valuable for sharing data across sites or for training models when data collection is incomplete. Table understanding tasks, including semantic parsing, table-to-text generation, and question answering over structured data, exploit the language understanding capabilities of models trained on both text and tables, and are increasingly relevant to clinical documentation and knowledge extraction workflows. Finally, representation learning treats the foundation model as a feature extractor, using its intermediate representations as inputs to lightweight classifiers, which enables efficient adaptation to new tasks without the computational cost of end-to-end retraining.


\section{Problem Statement}
\label{sec:problem}

Anyone who has tried to deploy a machine learning model in a clinical setting knows that choosing the right model is rarely straightforward. In the case of tabular foundation models, the problem is compounded by the sheer number of options now available, each developed with different priorities, different architectural choices, and different assumptions about the kind of data and tasks it will encounter. A model that performs excellently on a public benchmark may turn out to be unsuitable for a specific hospital's data because it cannot handle the way that hospital's schema is organized, or because it sends data to an external API that conflicts with patient privacy regulations, or because it requires more computational resources than the local infrastructure can provide. These are not abstract concerns; they are the kinds of issues that determine whether a model actually gets used in practice.

The core problem, then, is not a lack of capable models. It is the absence of a structured way to match available models to the requirements of a specific application. In healthcare, those requirements are particularly demanding. Privacy regulations such as GDPR and HIPAA impose hard constraints on how patient data can be handled, processed, and shared. Clinical datasets are often small or imbalanced, meaning a model that performs well on large benchmark datasets may fail when trained on data from a single department or a rare patient subgroup. Clinical decision support requires that predictions be interpretable and consistent, because a clinician who cannot understand why a model flagged a patient as high risk is unlikely to act on that flag, and may be right to be cautious. Fairness matters in a way that has direct consequences for patient outcomes, since a model that performs worse for certain demographic groups will systematically disadvantage those patients.

Other domains face their own version of this problem. In finance, the priorities shift toward computational speed and the ability to detect subtle patterns in high-dimensional transaction data. In manufacturing, scalability and the ability to handle evolving feature spaces may take precedence. The point is that the same foundation model cannot be the best choice everywhere, and making that selection thoughtfully requires a way to compare models not just on raw predictive accuracy but on the dimensions that actually matter for the deployment context.

A practical solution to this problem needs to do several things at once. It needs to cover a broad enough range of model properties that the most important factors in any given domain are accounted for. It needs to allow those properties to be weighted according to their relative importance in the specific context, because a high privacy score matters much more in a clinical trial registry than in an anonymized public health dataset. It needs to produce a ranking that is interpretable to the person making the selection decision, whether that person is a data scientist, a clinical informatics specialist, or a hospital administrator. And it needs to be maintainable, so that new models can be added to the comparison as the field advances.

There is a further requirement that is easy to overlook: the solution must be able to answer compound deployment questions by aggregating across evaluation dimensions, not just reporting them separately. Real deployment scenarios combine requirements from several dimensions at once, and practitioners need a single integrated ranking that reflects all of these priorities simultaneously. Single-dimension scores are useful for diagnosing a model's strengths and weaknesses, but the final selection decision requires a framework that combines them into a unified recommendation weighted by deployment priorities.

Formally, consider a set of available tabular foundation models $\mathcal{T}$, where each model $\tau_i$ has a set of characteristics $\mathcal{L}(\tau_i)$ that describe its properties. The goal is to identify the model that best satisfies the requirements of a given application domain $\Phi$, where each requirement $\phi_j$ in $\Phi$ carries a weight $w_{\phi_j}$ reflecting its importance in that domain. For each model, a score is computed by evaluating how well its characteristics meet each requirement and combining those scores into a single ranking value:

\begin{equation}
    \mathcal{R}(\tau_i) = \sum_{j=1}^m w_{\phi_j} \times Score(\mathcal{L}(\tau_i),\phi_j)
\end{equation}

where $Score(\mathcal{L}(\tau_i),\phi_j)$ measures how well model $\tau_i$ satisfies requirement $\phi_j$; in the framework described in the following section, each requirement $\phi_j$ corresponds to one sub-metric, and $Score(\mathcal{L}(\tau_i),\phi_j)$ is that sub-metric's score for $\tau_i$. The model selected for deployment is the one that maximizes this score:

\begin{equation}
    \tau_{best} \in \arg \max_{\tau_i \in \mathcal{T}} \mathcal{R}(\tau_i)
\end{equation}

When several models attain the same maximal score, all of them are equally recommended and the tie itself is informative, signaling that the deployment decision should be resolved with supplementary evidence, such as the super-metric scores described in Section~\ref{sec:supermetrics}.

This formulation makes the selection logic transparent: the final recommendation reflects explicit choices about what matters, and changing those choices, for instance by increasing the weight on privacy for a sensitive clinical dataset, changes the ranking in a predictable and auditable way. The framework we describe in the following section provides the structure to make this formulation operational.


\section{Design}
\label{sec:design}

Benchmark results alone do not tell a clinician or data scientist which tabular foundation model to deploy. Published results are typically obtained on different benchmark datasets selected according to the objectives and experimental settings of each study, and they rarely address the questions a deployment team actually needs answered: Will this model still generalize when applied to our hospital's EHR rather than a public benchmark dataset? Can it pass an information governance review? Will clinicians trust its explanations, and will its predictions be equally reliable for all patient groups? \system{} addresses this by organizing model evaluation around three layers, each answering a different kind of question, and by allowing the importance of each layer's components to be adjusted to reflect the priorities of the specific deployment context.

\subsection{Framework Architecture}
\label{sec:architecture}

\textbf{Sub-metrics} are the atomic units of evaluation. Each sub-metric is defined operationally: it corresponds to a specific, measurable aspect of model behavior that can, in principle, be quantified through an empirical test. Intra-Domain Transferability (IDT) is measured as the average performance drop when a model trained on one clinical dataset is evaluated on others from the same domain, such as blood test datasets from different laboratory networks. Membership Inference Resistance (MIR) is measured by the success rate of a membership inference attack against the trained model. Differential Privacy Compliance (DPC) is quantified by the formal privacy parameter $\epsilon$, where lower values represent stronger protection. Data Noise Robustness (DNR) is measured by the change in predictive performance as controlled levels of noise and missing data are introduced into the input. This operational grounding means that each sub-metric score is anchored, in principle, to a reproducible measurement rather than a subjective assessment.

In this paper, since direct empirical measurement of every sub-metric for every model would require running all models across dozens of clinical datasets, a task that is itself a substantial benchmarking contribution and is left as future work, sub-metric scores are derived from each model's documented design properties. A model that incorporates differential privacy mechanisms during training will predictably score higher on DPC than one that transmits raw patient data to an external inference API. A model with no mechanism for handling missing or additional features will score lower on Feature Adaptation Capability than one explicitly pre-trained on incomplete records. 
This property-based scoring approach replaces measured values with structured inference from documented design characteristics, while keeping every score traceable to the specific model properties that determined it.
The properties used and their effects on sub-metric scores are described in Section~\ref{sec:healthcare} and in Appendix~\ref{sec:appendix}.

\textbf{Metrics} aggregate sub-metrics within a clinical domain and produce a six-dimensional profile that describes each model's strengths and weaknesses in terms that practitioners can reason about. The six dimensions are Generalizability ($\mathcal{G}$), Privacy Preservation ($\mathcal{P}$), Data-Efficient Learning ($\mathcal{D}$), Scalability ($\mathcal{S}$), Interpretability ($\mathcal{I}$), and Fairness and Bias Mitigation ($\mathcal{F}$). Within each dimension, sub-metric scores are combined through a weighted sum, where the weights are set by the practitioner to reflect the priorities of the deployment context. A clinical system handling identifiable patient data will assign higher weights to privacy sub-metrics than a research study working with fully anonymized records. This means the framework produces different, and more informative, model profiles for different deployment scenarios without requiring a different evaluation system for each one.

\textbf{Super-metrics} are supplementary compound scores that combine sub-metrics from several metric domains to provide a diagnostic view of how models perform across specific cross-domain concerns. They do not replace $\mathcal{R}$ as the basis for model selection; they help practitioners understand the strengths and trade-offs of top-ranked models in greater depth. The four super-metrics are described in Section~\ref{sec:supermetrics} below. Figure~\ref{fig:architecture} illustrates the complete three-layer architecture and the flow from individual sub-metrics up through the overall ranking.

\begin{figure}[t]
    \centering
    \includegraphics[width=\linewidth]{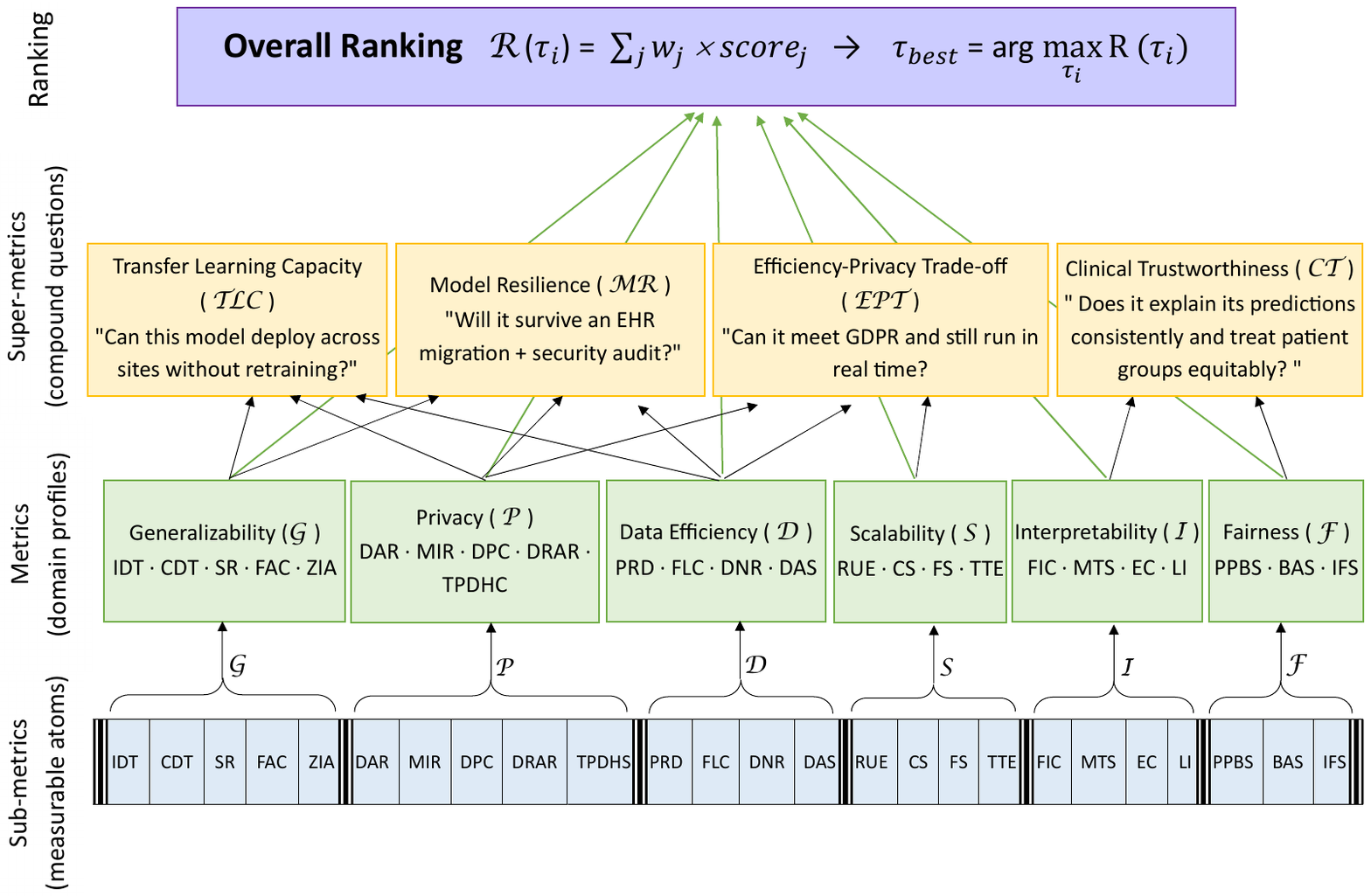}
    \caption{The three-layer evaluation architecture of \system{}. Sub-metrics (bottom, grouped by metric domain) are atomic measurable properties derived from documented model design characteristics. Metrics (middle) aggregate sub-metrics within six clinical evaluation dimensions using domain-specific weights. Super-metrics (top) are optional supplementary compound scores that combine sub-metrics across domain boundaries to provide additional diagnostic depth. The overall ranking $\mathcal{R}$ integrates all sub-metric scores into a single recommended model for the target deployment context.}
    \label{fig:architecture}
\end{figure}

\subsection{Comparison Metrics}

\subsubsection{Generalizability}
In clinical informatics, the same kind of data rarely looks identical from one institution to another. A dataset from a university hospital may include dozens of laboratory features that a regional clinic never collects. Column names differ between systems, values are encoded differently, and some features that are routinely available in one setting are simply absent in another. A model that only works well when the data matches its training set exactly has limited practical value. Generalizability measures how well a model adapts when the data it encounters is different from what it was trained on.

Intra-Domain Transferability (IDT) captures how consistently a model performs across different datasets within the same clinical domain, for example across blood test datasets from different institutions. A model with strong IDT can be deployed at a new hospital without retraining from scratch, simply because it has learned representations general enough to transfer within the domain. Cross-Domain Transferability (CDT) extends this further, measuring how well the model handles data from related but distinct domains, such as moving from clinical to administrative healthcare data. Schema Robustness (SR) focuses specifically on structural variation: does the model still work if columns are reordered, if a single variable is split into two, or if feature names differ between datasets? Feature Adaptation Capability (FAC) asks what happens when key features are missing or when new, previously unseen features are introduced. Zero-Shot Inference Adaptability (ZIA) measures the most demanding form of generalization, performance on a completely new dataset without any fine-tuning at all.

The overall generalizability score ($\mathcal{G}$) is:

\begin{eqnarray*}
    \mathcal{G}  & = & \sum_{i=1}^{|S_g|} (w_i \times submetric_i) \\
    s.t. & & S_g = \{ \text{IDT, CDT, SR, FAC, ZIA} \}
\end{eqnarray*}

\subsubsection{Privacy Preservation}
Patient data is among the most sensitive information that exists. In most jurisdictions, its collection, storage, and use are governed by regulations such as GDPR and HIPAA that impose strict requirements on how data can be processed and shared. A tabular foundation model that achieves excellent predictive accuracy but exposes patient information through its outputs or training process is not deployable in a real clinical setting. Privacy preservation measures how well a model holds up against the range of attacks and compliance requirements that are relevant in practice.

Data Anonymization Robustness (DAR) measures whether the model still performs well when trained on anonymized data. Clinical datasets are often anonymized before they can be shared across institutions, and a model that loses significant accuracy under anonymization is harder to use in collaborative research settings. Membership Inference Resistance (MIR) addresses a specific and well-documented attack: an adversary who tries to determine whether a particular patient's record was part of the training set. Models that memorize training examples rather than learning generalizable patterns are vulnerable to this attack, which is a direct privacy risk for the individuals in the training data. Differential Privacy Compliance (DPC) measures how well the model meets the formal standards of differential privacy, quantified by the privacy parameter $\epsilon$, where lower values represent stronger protection. Data Reconstruction Attack Resistance (DRAR) measures how difficult it is for an attacker to reconstruct original records from the model's outputs or gradients, a threat that is particularly relevant for generative models that can produce synthetic data. Third-Party Data Handling Compliance (TPDHC) evaluates whether the model's design and deployment respect privacy requirements when data is shared with or processed by external services, covering practices such as encryption, access control, and consent management.

The overall privacy preservation score ($\mathcal{P}$) is:

\begin{eqnarray*}
    \mathcal{P}  & = & \sum_{i=1}^{|S_p|} (w_i \times submetric_i) \\
    s.t. & & S_p = \{ \text{DAR, MIR, DPC, DRAR, TPDHC} \}
\end{eqnarray*}

\subsubsection{Data-Efficient Learning}
One of the most persistent challenges in clinical AI is the scarcity of labeled data. Annotating patient records requires clinical expertise, which is expensive and time-consuming. Some disease areas have small patient populations by definition, and historical datasets may not cover the range of presentations that a model needs to learn. A model that requires tens of thousands of labeled examples to perform well is simply not viable in many healthcare contexts. Data-efficient learning captures how much a model can achieve with limited data.

Performance with Reduced Data (PRD) measures how well the model maintains its accuracy as the training set shrinks. This directly reflects the practical scenario where a hospital wants to deploy a model trained on its own patient population, which may be much smaller than the large public datasets models are typically benchmarked on. Few-Shot Learning Capability (FLC) goes further, measuring performance when only a handful of labeled examples per class are available, a common scenario for rare conditions or underrepresented patient subgroups. Data Noise Robustness (DNR) measures how the model handles the kind of imperfect data that is routine in clinical practice: missing laboratory values, inconsistent coding, data entry errors, and sensor noise. Data Augmentation Sensitivity (DAS) captures how effectively the model uses synthetic data to compensate for limited real examples, which is relevant when augmentation is part of the training pipeline.

The overall data-efficient learning score ($\mathcal{D}$) is:

\begin{eqnarray*}
    \mathcal{D}  & = & \sum_{i=1}^{|S_d|} (w_i \times submetric_i) \\
    s.t. & & S_d = \{ \text{PRD, FLC, DNR, DAS} \}
\end{eqnarray*}

\subsubsection{Scalability}
Clinical datasets vary enormously in size, from a few hundred patients in a single-center retrospective study to millions of records in a national registry. Feature spaces vary just as much, from a focused set of a dozen laboratory values to comprehensive EHR extracts with hundreds of variables. A model that works well at one scale may fail at another, either because it becomes too slow to train, too memory-intensive to deploy, or because its performance degrades as the data grows more complex. Scalability measures how gracefully a model handles these transitions.

Resource Utilization Efficiency (RUE) compares the ratio of performance gain to resource cost as dataset size increases, capturing whether the model uses memory and compute proportionally or disproportionately as it scales. Computational Scalability (CS) focuses specifically on whether computational demand grows at a manageable rate when more data is added. Feature Scalability (FS) measures performance as the number of features increases, which is important in EHR-derived datasets where the feature space can expand substantially depending on which data sources are included. Training Time Efficiency (TTE) evaluates whether additional training time continues to yield meaningful improvements, or whether the model plateaus quickly, which has practical implications for iterative deployment and model updating workflows.

The overall scalability score ($\mathcal{S}$) is:

\begin{eqnarray*}
    \mathcal{S}  & = & \sum_{i=1}^{|S_s|} (w_i \times submetric_i) \\
    s.t. & & S_s = \{ \text{RUE, CS, FS, TTE} \}
\end{eqnarray*}

\subsubsection{Interpretability}
A prediction that clinicians cannot understand is a prediction they are unlikely to trust, and for good reason. Clinical decisions carry consequences for patients, and the responsibility for those decisions rests with the clinician, not the model. For a model to function as a useful clinical decision support tool, it needs to be able to communicate not just what it predicted but why, in terms that a clinician can evaluate against their own clinical knowledge. Interpretability measures how well a model supports this kind of human oversight.

Feature Importance Clarity (FIC) measures how well the model identifies which features drove a given prediction, and whether those identified features align with what clinical experts would expect. A model that identifies hemoglobin and ferritin as the most important features for an iron deficiency prediction is more interpretable, and more trustworthy, than one that assigns high importance to a feature with no clear clinical rationale. Model Transparency Score (MTS) reflects the overall comprehensibility of the model's structure, accounting for factors like the number of parameters, layers, and attention heads relative to any built-in interpretability mechanisms. Explanation Consistency (EC) measures whether the model provides similar explanations for similar patients, which matters because inconsistent explanations undermine the confidence of clinicians who are trying to understand the model's reasoning. Local Interpretability (LI) evaluates the quality of patient-level explanations generated by post-hoc methods such as SHAP values or LIME, measuring how accurately those explanations represent the model's actual behavior for individual predictions.

The overall interpretability score ($\mathcal{I}$) is:

\begin{eqnarray*}
    \mathcal{I}  & = & \sum_{j=1}^{|S_i|} (w_j \times submetric_j) \\
    s.t. & & S_i = \{ \text{FIC, MTS, EC, LI} \}
\end{eqnarray*}

\subsubsection{Fairness and Bias Mitigation}
Healthcare data reflects the world as it has been, not as it should be. Historical datasets encode patterns of access, documentation, and diagnosis that are shaped by systemic inequalities. A model trained on such data without correction can reproduce and even amplify those inequalities, producing predictions that are less accurate for women, for elderly patients, for ethnic minorities, or for other groups that have been underrepresented or differently treated in the training data. In a clinical context, this means that the patients who already face the greatest barriers to care may be the ones for whom the model performs worst. Fairness and bias mitigation measures how effectively a model avoids this outcome.

Predictive Parity Bias Score (PPBS) measures whether the positive predictive value of the model is consistent across demographic groups. A model with high PPBS is one where, if it predicts a patient is at high risk, that prediction is equally reliable regardless of whether the patient is young or old, male or female, or from a majority or minority population. Bias Amplification Score (BAS) measures whether the model makes existing biases in the training data worse or keeps them roughly constant. Intersectional Fairness Score (IFS) addresses the more complex case where multiple demographic factors interact, for instance where the performance gap affects older women specifically, rather than older patients or women as separate groups. A higher IFS indicates the model performs equitably across these intersectional subgroups.

The overall fairness and bias mitigation score ($\mathcal{F}$) is:

\begin{eqnarray*}
    \mathcal{F}  & = & \sum_{i=1}^{|S_f|} (w_i \times submetric_i) \\
    s.t. & & S_f = \{ \text{PPBS, BAS, IFS} \}
\end{eqnarray*}

While this paper demonstrates the framework through healthcare, the same three-layer architecture applies to any domain where tabular data drives high-stakes decisions. In finance, scalability and fraud-detection metrics take priority; in manufacturing, evolving sensor feature spaces shift weight toward adaptability; in public administration, fairness metrics become first-order concerns. The metric weights are the mechanism through which these domain-specific priorities translate into differentiated model rankings without requiring a different evaluation system for each context.

\subsection{Super-Metrics}
\label{sec:supermetrics}

The six metrics described above each capture one dimension of model behavior, and the overall ranking $\mathcal{R}(\tau_i)$, the weighted sum across all sub-metrics, is the primary output of the framework and the basis for model selection. Alongside this ranking, the framework provides four supplementary compound scores called super-metrics. Each super-metric combines sub-metrics from several metric domains to offer a diagnostic view of how a model performs under a specific cross-domain concern. Super-metrics do not replace $\mathcal{R}$: they help practitioners understand the strengths and trade-offs of top-ranked models in greater depth. A practitioner would use $\mathcal{R}$ to identify the recommended model and use the super-metric scores to understand its profile across compound concerns relevant to their context.

\begin{figure}
	\centering
	\begin{subfigure}{0.24\linewidth}
		\includegraphics[width=\linewidth]{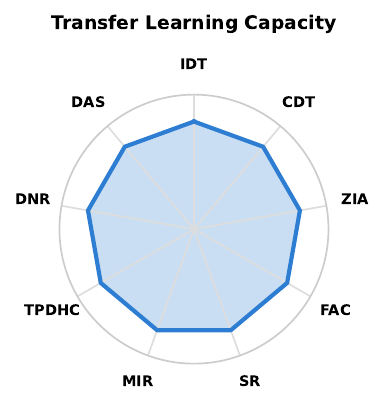}
		\caption{Transfer Learning Capacity}
		\label{fig:transfer}
	\end{subfigure}
	\begin{subfigure}{0.24\linewidth}
		\includegraphics[width=\linewidth]{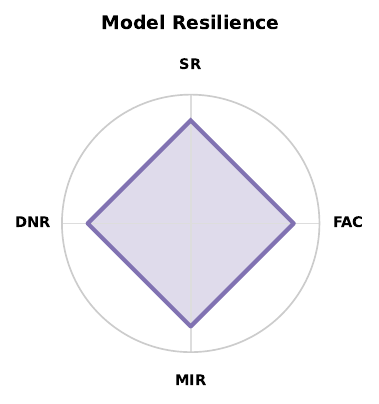}
		\caption{Model Resilience}
		\label{fig:resilience}
	\end{subfigure}
	\begin{subfigure}{0.24\linewidth}
		\includegraphics[width=\linewidth]{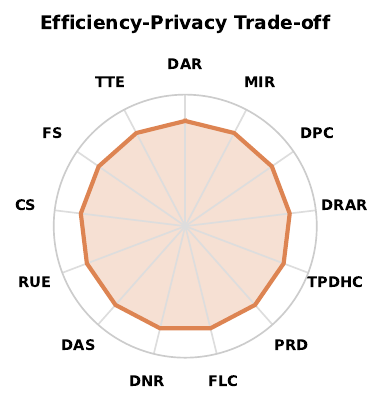}
		\caption{Efficiency-Privacy Trade-off}
		\label{fig:tradeoff}
	\end{subfigure}
	\begin{subfigure}{0.24\linewidth}
		\includegraphics[width=\linewidth]{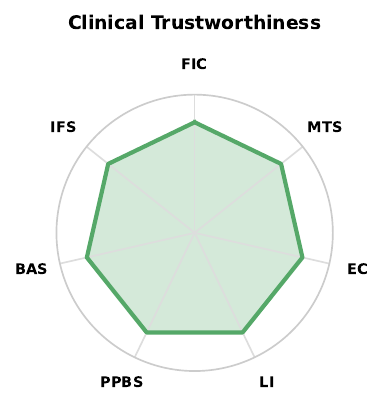}
		\caption{Clinical Trustworthiness}
		\label{fig:ct}
	\end{subfigure}
	\caption{Sub-metric composition of the four supplementary super-metrics, shown as radar plots at reference maximum values. TLC draws from G, P, and D; MR from G, P, and D; EPT from P, D, and S; CT from I and F. Together they cover all six metric domains. When actual model scores replace the reference values, the polygon shape makes cross-domain strengths and deficits visible as a diagnostic complement to the overall ranking $\mathcal{R}$.}
	\label{fig:super-metrics}
\end{figure}

\textbf{Transfer Learning Capacity} ($\mathcal{TLC}$) captures how well a model supports safe deployment across multiple institutions simultaneously. Cross-institutional transfer requires not only that the model generalizes across schemas and domains, but also that it resists privacy attacks during transfer and handles data quality variation at new sites without retraining. TLC combines all five generalizability sub-metrics with two privacy sub-metrics (MIR, TPDHC) and two data-efficiency sub-metrics (DNR, DAS), the sub-metrics that must hold together for multi-site deployment to be both practical and safe (see Figure~\ref{fig:transfer}):

\begin{eqnarray*}
    \mathcal{TLC} & = & \sum_{i=1}^{|S_{tlc}|} (w_i \times submetric_i) \\
    s.t. & & S_{tlc} = \{ \text{IDT, CDT, ZIA, FAC, SR, MIR, TPDHC, DNR, DAS} \}
\end{eqnarray*}

\textbf{Model Resilience} ($\mathcal{MR}$) captures a model's ability to remain functional under real-world operational pressures that arise simultaneously: structural changes from EHR migrations or schema updates (SR, FAC), privacy threats during transition periods when external parties have model access (MIR), and elevated data noise from inconsistent coding or data quality degradation (DNR). These stresses co-occur in practice, and MR scores them together as a single diagnostic view of operational robustness (see Figure~\ref{fig:resilience}):

\begin{eqnarray*}
    \mathcal{MR} & = & \sum_{i=1}^{|S_{mr}|} (w_i \times submetric_i) \\
    s.t. & & S_{mr} = \{ \text{SR, FAC, MIR, DNR} \}
\end{eqnarray*}

\textbf{Efficiency-Privacy Trade-off} ($\mathcal{EPT}$) addresses a three-way tension that is common in hospital deployments: regulatory privacy compliance, data-efficient operation under limited training data, and computational scalability within hospital IT infrastructure. Each of these is captured by a separate metric domain, and their sub-metrics already contribute individually to $\mathcal{R}$. EPT presents them in aggregate as a single cross-domain diagnostic view of this three-way constraint, making simultaneous trade-offs immediately visible rather than requiring inspection of three separate metric scores (see Figure~\ref{fig:tradeoff}):

\begin{eqnarray*}
    \mathcal{EPT} & = & \sum_{i=1}^{|S_{ept}|} (w_i \times submetric_i) \\
    s.t. & & S_{ept} = \{ \text{DAR, MIR, DPC, DRAR, TPDHC, PRD, FLC, DNR, DAS, RUE, CS, FS, TTE} \}
\end{eqnarray*}

\textbf{Clinical Trustworthiness} ($\mathcal{CT}$) captures the two properties that clinical ethics committees and governance bodies most commonly require before approving a clinical AI system: the model must be explainable to the clinicians who use it, and it must perform equitably across all patient groups. Interpretability and fairness are evaluated separately in the metric layer, but they must hold together in practice: a model that explains itself well but produces biased predictions is not approvable, and a model that is fair but opaque will not be adopted. CT combines all four interpretability sub-metrics with all three fairness sub-metrics as a single diagnostic view of clinical governance readiness (see Figure~\ref{fig:ct}):

\begin{eqnarray*}
    \mathcal{CT} & = & \sum_{i=1}^{|S_{ct}|} (w_i \times submetric_i) \\
    s.t. & & S_{ct} = \{ \text{FIC, MTS, EC, LI, PPBS, BAS, IFS} \}
\end{eqnarray*}

Because all four super-metric scores use the same sub-metric weights as the individual metrics, they automatically inherit the deployment priorities of the context. A privacy-sensitive deployment will produce a more demanding TLC and MR score than a research setting that relaxes privacy weights; a deployment in a context with documented demographic disparities will produce a more demanding CT score. Together the four super-metrics span all six metric domains, so every dimension of model behavior contributes to at least one compound diagnostic view.

\section{Healthcare: a case study}
\label{sec:healthcare}

Healthcare is in many ways the most demanding domain for tabular foundation models. The data is sensitive, the decisions are high-stakes, the regulatory environment is strict, and the labeled datasets available for training are often much smaller than what models in other domains are trained on. At the same time, the potential benefits are substantial. Clinical databases contain decades of patient records, laboratory measurements, diagnoses, prescriptions, and outcomes that, if analyzed well, could support earlier diagnosis, more accurate risk stratification, and more personalized treatment. Tabular foundation models are well placed to extract that value, because the structured nature of clinical data is exactly what these models are designed to handle.

Applying \system{} to healthcare means working through the three-layer architecture in a clinical context. The first step is to establish which of the six metrics matter most for the specific task, because the same six dimensions are not equally important in every deployment. The second step is to translate those metric-level priorities into sub-metric weights within each dimension, since the sub-metrics are what connect the clinical requirements to measurable model properties. The third step is to read the outputs of the framework: a metric profile that shows each model's strengths and weaknesses dimension by dimension, supplementary super-metric scores that provide additional diagnostic depth on cross-domain concerns, and the overall ranking $\mathcal{R}(\tau_i)$ that integrates everything into a single recommendation. We demonstrate this full process through two use cases that differ substantially in their clinical context, their data characteristics, and their deployment requirements, and we show how those differences propagate through all three layers of the framework.

\subsection{Tabular Foundation Models Characteristics}

Before applying the framework to a specific use case, the first step is to characterize the models being compared. Rather than relying solely on published benchmark results, which are often collected on datasets quite different from clinical ones, we score each model against a set of qualitative properties that reflect its design and capabilities. Each property either positively or negatively influences a set of relevant sub-metrics. Models start with a default sub-metric score of 10, and each missing or problematic property reduces the relevant sub-metric scores by 1. The properties and their effects are summarized in Table~\ref{tab:drawback_properties}.

It is important to note that a score of 10 does not indicate empirically demonstrated perfect performance on a sub-metric. Rather, it indicates that no design-property limitation associated with that sub-metric was identified based on the available evidence. The resulting scores should therefore be
interpreted as property-based estimates of model suitability for comparison within the framework, rather than as substitutes for direct empirical measurement.

\begin{table}[H]
\centering
\small
\begin{tabular}{|p{3cm}|p{8cm}|p{3cm}|}
\hline
\textbf{Property} & \textbf{Brief Description} & \textbf{Sub-metrics Decreased by 1 (if absent)}\\
\hline
Built-in Pre-Training &
The model has a genuine pre-training phase (self-supervised pretraining on auxiliary data, or a meta-learned/prior-fitted setup used before task-specific application), increasing its adaptability and performance in low-data and zero-shot scenarios. End-to-end supervised training on the target dataset alone does not satisfy this property. &
IDT, CDT, FAC, ZIA, PRD, FLC \\
\hline
On-Premises Data Processing &
The model processes all data locally without sending raw or sensitive records to external services, preserving privacy and regulatory compliance. &
DAR, MIR, DPC, DRAR, TPDHC \\
\hline
Handle Missing Data &
The model can gracefully handle incomplete records, increasing robustness to schema and noise variations. &
SR, DNR \\
\hline
Explainability Tools &
The model provides interpretable outputs or feature importance, supporting transparency and trust. &
FIC, MTS, EC, LI \\
\hline
Predictive Fairness Constraints&
The model incorporates mechanisms to equalize positive predictive value across demographic groups, reducing disparate impact in high-risk predictions. &
PPBS, IFS \\
\hline
Bias Amplification Control&
The model does not amplify biases present in the training data; its predictions are no more biased across groups than the training data itself. &
BAS \\
\hline
Handle Noisy Inputs &
The model is not easily thrown off by noisy or perturbed data, increasing its reliability in real-world settings. &
DNR \\
\hline
Embedding Efficiency &
The model does not generate excessively large embeddings, or consume excessive memory and computational resources.  &
FS, RUE, CS, TTE \\
\hline
Accelerated Embedding Generation &
The process of creating embeddings is fast enough or resource-efficient, increasing overall computational efficiency. &
RUE, CS, TTE \\
\hline
Data Anonymization Techniques &
The model anonymizes or protects sensitive features, decreasing vulnerability to privacy attacks and improving compliance when data must be shared with third parties. &
DAR, MIR, DPC, DRAR, TPDHC \\
\hline
Flexibility to Evolving Schemas &
The model is able to adapt to changes in table structure (e.g., new or missing columns), increasing robustness. &
SR \\
\hline
Scaling with Growing Features &
The model adapts to handle an increasing number of features, enhancing its capability to manage complex datasets. &
FS \\
\hline
Data Augmentation Capability &
The model leverages augmented data to improve robustness and performance, enabling generalization in low-data conditions. &
DAS, PRD, FLC \\
\hline
Efficient Resource Utilization &
The model optimizes computation and memory usage to ensure efficiency. &
RUE, CS, TTE \\
\hline
\end{tabular}
\caption{Properties for Tabular Foundation Models. The absence of each property decreases the corresponding sub-metrics by 1, starting from a default value of 10. Built-in Pre-Training and Data Anonymization Techniques each now govern one additional sub-metric (ZIA and TPDHC respectively) relative to the original mapping, correcting two gaps under which those sub-metrics could not be reduced below the maximum for any model regardless of documented capability.}
\label{tab:drawback_properties}
\end{table}

An earlier version of this scoring pass applied this rule directly only to a subset of models and relied on holistic architectural judgment for the remainder, which allowed a small cluster of models to be scored near the maximum on properties their own publications do not document (for example, crediting a purely end-to-end supervised architecture with a pretraining-linked property it explicitly lacks). We subsequently re-derived the property assignment for every model in the survey against its original publication, applying the rule in Table~\ref{tab:drawback_properties} mechanically and uniformly, with no per-model exceptions. Where a paper's evidence for a property was ambiguous rather than clearly documented, the property is scored absent, consistent with the traceability requirement above. Two gaps in the property-to-sub-metric mapping surfaced during this re-derivation and are corrected in Table~\ref{tab:drawback_properties}: Zero-Shot Inference Adaptability (ZIA) was not linked to any property, meaning it could not be reduced below the maximum for any model regardless of its documented capabilities; it is now linked to Built-in Pre-Training, since zero-shot inference without task-specific fine-tuning is a direct consequence of genuine pretraining or prior-fitting. Third-Party Data Handling Compliance (TPDHC) was linked only to On-Premises Data Processing; it is now also linked to Data Anonymization Techniques, consistent with how the other privacy sub-metrics are already jointly governed by both properties. The re-derivation also surfaced four entries in the extended model list whose original identification could not be substantiated against primary literature under their listed name (P-Transformer, DTT, GPT4Table, and SPROUT/UniTTab, detailed in Appendix~\ref{sec:appendix}); these are flagged rather than silently scored, and we recommend they be replaced or independently re-verified in any subsequent revision.

\begin{figure}[t]
    \centering
    \includegraphics[width=\linewidth]{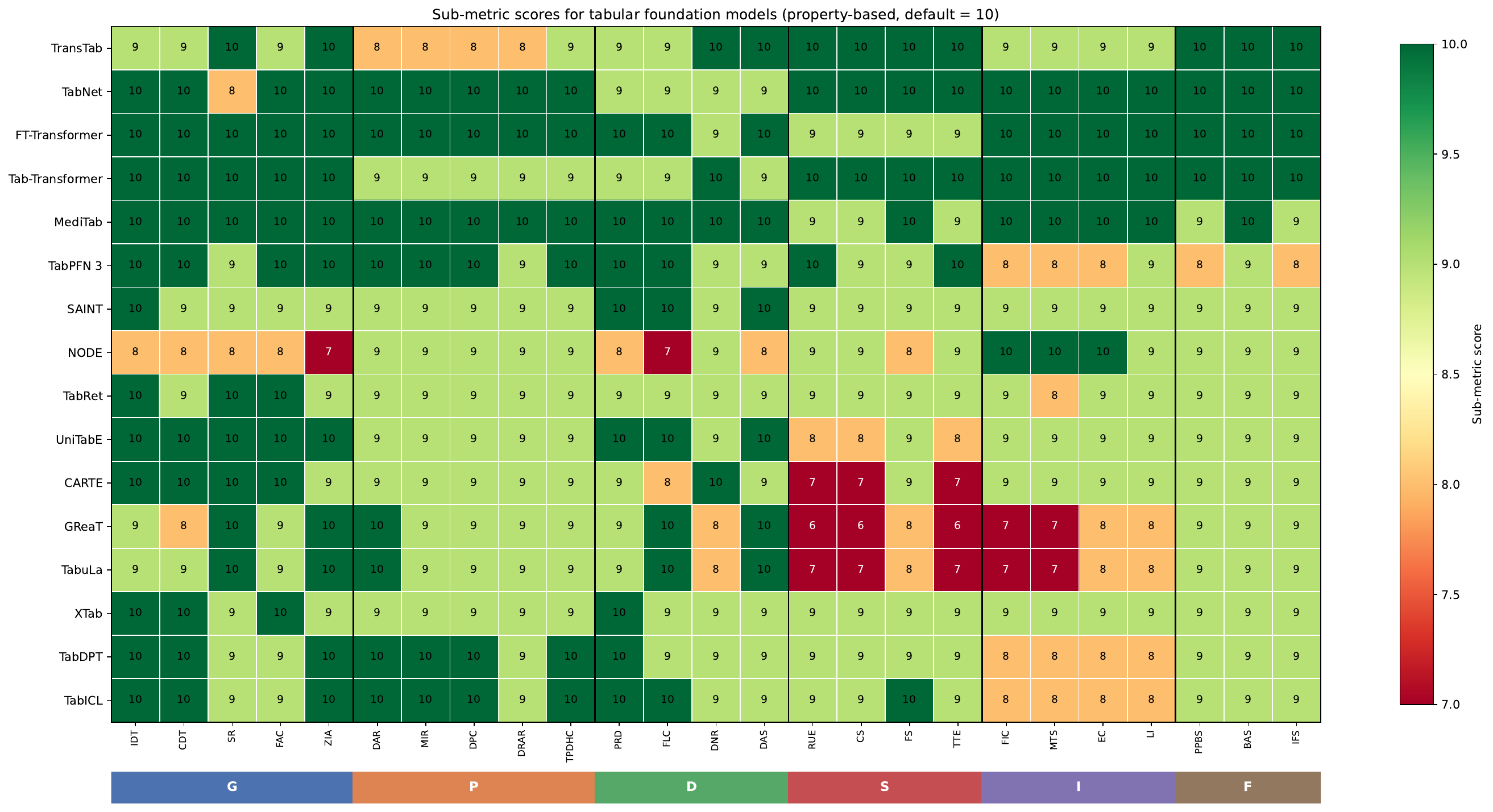}
    \caption{Sub-metric scores for 16 tabular foundation models across all 25 sub-metrics, derived from documented design properties starting from a default of 10. Columns are grouped by metric domain (G\,=\,Generalizability, P\,=\,Privacy, D\,=\,Data Efficiency, S\,=\,Scalability, I\,=\,Interpretability, F\,=\,Fairness). Green cells indicate maximum scores; yellow and orange indicate single- and double-point deductions; red indicates three or more points below the maximum. Models are ordered from highest to lowest aggregate score. The full property-to-sub-metric mapping is given in Table~\ref{tab:drawback_properties}.}
    \label{fig:tfm_scores}
\end{figure}

This approach makes the scoring process transparent and auditable. A clinician or data scientist reviewing a model ranking can trace any score back to specific design properties of the model, understand why a model scored lower on privacy or interpretability, and decide whether those trade-offs are acceptable in their specific deployment context. Figure~\ref{fig:tfm_scores} presents the resulting sub-metric scores for the 45 tabular foundation models surveyed in this work across all 25 sub-metrics, computed in this way from their reported design properties. These scores were subsequently re-derived a second time against independent comparison and survey literature rather than each model's own publication alone (Section~\ref{sec:audit-note}). The eight models evaluated in the case study ranking tables (Section~\ref{sec:id-usecase}, Section~\ref{sec:hf-usecase}) are the eight with the strongest combined ranking across both use cases under this independent-literature-based pass: TabPFN~2.5 and TabICL, tied for the strongest combined ranking, followed by VIME, MediTab, TAPTAP, TabDPT, SAINT, and XTab. This is not the same set of eight models identified by the first, primary-source-only audit pass (TabPFN~v2, TabICL, TabPFN~2.5, SAINT, VIME, TabPFN, TabNet, and MediTab): TabPFN~v2, TabPFN, and TabNet no longer place in the top eight once each model's true peer set, rather than its own publication, is taken into account, and TAPTAP, TabDPT, and XTab take their place. As before, this set is not chosen in advance to illustrate a particular narrative; it is simply where the independent-literature-based scores place the strongest candidates, and the reasons the ranking changed are themselves part of the finding (Section~\ref{sec:audit-note}).

\subsection{Iron Deficiency Use Case}
\label{sec:id-usecase}

Iron deficiency is the most prevalent nutritional deficiency worldwide, yet it remains chronically underdiagnosed. A particular challenge is iron deficiency without anemia (IDWA), the stage at which iron stores are depleted but hemoglobin has not yet fallen below the clinical threshold. Patients at this stage are symptomatic, with fatigue, reduced exercise tolerance, and impaired cognitive function, but standard hemoglobin-based screening misses them entirely. Ferritin testing can detect IDWA, but conventional ferritin thresholds are poorly calibrated across sex, age, and inflammatory status, leading to large numbers of false negatives in routine clinical practice.

Machine learning has emerged as a promising approach to close this diagnostic gap. Efros and colleagues demonstrated that an extreme gradient-boosted model (XGBoost) trained on routine EHR data from 211,486 patients across the Mount Sinai Health System could predict low ferritin levels in non-anemic patients with an AUC of 0.814, meaningfully outperforming a model based on complete blood count indices alone \cite{efros2025predictive}. Importantly, the model's performance varied by demographic subgroup, with lower accuracy in premenopausal women, a finding that directly motivates fairness-aware evaluation in any deployed system. In a parallel study using CBC and reticulocyte maturation indices from a hospital cohort, Farhanpoor and colleagues found that a CatBoost classifier achieved an AUC of 0.98 in cross-validation and 0.93 on an independent test set, with SHAP analysis identifying hemoglobin, hematocrit, and red blood cell count as the most influential predictors, which aligns with the clinical understanding of iron-deficient erythropoiesis \cite{farhanpoor2025ai}. Earlier work using deep learning on longitudinal outpatient laboratory sequences showed that it is possible to identify patients at risk of developing iron deficiency anemia substantially ahead of the clinical diagnosis, offering a window for preventive intervention \cite{garduno2024early}. Taken together, these studies establish that the tabular, laboratory-derived data used in routine hematology practice contains enough signal to support effective machine learning screening, and that the challenge is not whether such models can work, but which model architecture and evaluation framework is most appropriate for a given clinical deployment.

Applying \system{} to this context begins at the metric level. Privacy Preservation and Interpretability are the two highest-priority dimensions: patient laboratory data is protected under GDPR and HIPAA, and the demographic variability in iron deficiency risk means fairness-unaware predictions could actively harm the most vulnerable patients. Generalizability ranks equally high, because any practically useful screening program must work across institutions without per-site retraining. Data-Efficient Learning is elevated above Scalability, reflecting the reality that single-site hematology datasets are typically small. Scalability is the lowest-priority metric in this context, since iron deficiency screening does not require the scale of processing that characterizes national registries or real-time transaction systems. These metric-level priorities then determine the sub-metric weights within each dimension.

Under the Generalizability metric, Intra-Domain Transferability (IDT) receives a weight of 5, because blood test datasets from different clinical laboratories and institutions vary in their reference ranges, coding conventions, and included analytes, and a model that cannot transfer across these variations will need to be retrained separately for each site. Cross-Domain Transferability (CDT) receives a weight of 2, reflecting the fact that iron deficiency detection is clinically specific enough that generalization to unrelated domains adds little practical value. Schema Robustness (SR) receives a weight of 4, because healthcare records frequently present the same clinical information in different structural formats, and the model must handle schema variations without failing. Feature Adaptation Capability (FAC) receives the highest weight of 5, given how frequently clinical datasets have missing or additional variables, for instance when a laboratory panel is expanded or when data is collected from a setting that does not routinely measure a particular analyte. Zero-Shot Inference Adaptability (ZIA) receives a weight of 3, reflecting that while zero-shot capability is useful, the availability of at least some domain-specific data for fine-tuning reduces its priority in this context.

Under the Privacy Preservation metric, all sub-metrics are weighted heavily, reflecting the sensitivity of patient laboratory data. Data Anonymization Robustness (DAR) receives a weight of 5, because clinical datasets for iron deficiency screening routinely require de-identification before they can be used or shared, and a model that degrades substantially under anonymization is less useful for collaborative or multi-site deployment. Membership Inference Resistance (MIR) receives a weight of 4, as the ability to infer whether a specific patient's data was in the training set is a direct privacy risk. Differential Privacy Compliance (DPC) receives a weight of 4, reflecting the regulatory environment governing clinical data in most jurisdictions. Data Reconstruction Attack Resistance (DRAR) receives a weight of 5, given the sensitivity of laboratory values as identifiers. Third-Party Data Handling Compliance (TPDHC) receives a weight of 3, which is moderate rather than high because the screening scenario we consider here involves primarily institution-internal deployment rather than extensive third-party data sharing.

Under the Data-Efficient Learning metric, Performance with Reduced Data (PRD) and Few-Shot Learning Capability (FLC) both receive a weight of 5. Iron deficiency screening datasets from individual clinical sites are typically limited in size, and the subpopulations of greatest clinical interest, pregnant women, children, patients with inflammatory conditions, are often underrepresented. A model that can learn from limited labeled data is a model that can be realistically deployed across a range of clinical settings. Data Noise Robustness (DNR) receives a weight of 4, since laboratory data quality varies across collection sites and time periods. Data Augmentation Sensitivity (DAS) receives a weight of 4, as synthetic augmentation is a practical tool for addressing data scarcity, though it is secondary to learning from real patient data.

Under the Scalability metric, the weights are lower overall. Resource Utilization Efficiency (RUE) receives a weight of 3, reflecting moderate importance in resource-constrained environments. Computational Scalability (CS) receives a weight of 2, since population-level iron deficiency screening is not typically performed at a scale that would stress most model architectures. Feature Scalability (FS) receives a weight of 4, because laboratory panels for iron deficiency can include a range of analytes, and the model must handle high-dimensional feature sets without degradation. Training Time Efficiency (TTE) receives a weight of 2, as accuracy and reliability take priority over training speed in this clinical context.

Under the Interpretability metric, all sub-metrics are highly weighted. Feature Importance Clarity (FIC) receives a weight of 5, because clinicians need to understand which laboratory values are driving a prediction in order to evaluate it against their clinical knowledge and to trust the model's output. Model Transparency Score (MTS) receives a weight of 4, reflecting the importance of model comprehensibility for clinical adoption. Explanation Consistency (EC) receives a weight of 5, since inconsistent explanations for similar patients would quickly undermine clinician trust. Local Interpretability (LI) receives a weight of 5, because patient-level explanations are central to clinical decision support: the clinician needs to understand not just what the model predicts on average, but why it is flagging this particular patient.

Under the Fairness and Bias Mitigation metric, Predictive Parity Bias Score (PPBS) receives a weight of 5, given the well-documented demographic disparities in iron deficiency prevalence and diagnosis, particularly across age groups and pregnancy status. Bias Amplification Score (BAS) receives a weight of 4, emphasizing the importance of ensuring that the model does not worsen existing disparities in the training data. Intersectional Fairness Score (IFS) receives a weight of 5, reflecting that intersecting demographic factors such as age and sex are directly clinically relevant to iron deficiency risk.

\begin{figure}[t]
\centering
\begin{tikzpicture}[x=1cm, y=1cm, font=\sffamily]

\node[font=\normalsize\bfseries\sffamily, anchor=south]
  at ({7.500}, 0.20) {Iron Deficiency};

\fill[mG] (3.600,0) rectangle (5.500,-0.400);
\node[text=white,font=\small\bfseries\sffamily] at (4.550,-0.200) {G};
\fill[mP] (5.500,0) rectangle (7.400,-0.400);
\node[text=white,font=\small\bfseries\sffamily] at (6.450,-0.200) {P};
\fill[mD] (7.400,0) rectangle (9.300,-0.400);
\node[text=white,font=\small\bfseries\sffamily] at (8.350,-0.200) {D};
\fill[mS] (9.300,0) rectangle (11.200,-0.400);
\node[text=white,font=\small\bfseries\sffamily] at (10.250,-0.200) {S};
\fill[mI] (11.200,0) rectangle (13.100,-0.400);
\node[text=white,font=\small\bfseries\sffamily] at (12.150,-0.200) {I};
\fill[mF] (13.100,0) rectangle (15.000,-0.400);
\node[text=white,font=\small\bfseries\sffamily] at (14.050,-0.200) {F};
\fill[black!15] (0,0) rectangle (3.600,-0.400);
\node[anchor=west,font=\small\bfseries\sffamily] at (0.10,-0.200) {Model};
\node[anchor=west,font=\small\bfseries\sffamily] at (15.150,-0.200) {$\mathcal{R}$};

\fill[black!5] (0,-0.400) rectangle (3.600,-1.280);
\node[anchor=west,font=\small\sffamily,inner sep=2pt] at (0.10,-0.840) {TabPFN~2.5};
\fill[fill={rgb,255:red,57;green,167;blue,88}] (3.600,-0.400) rectangle (5.500,-1.280);
\node[text=white,align=center,font=\scriptsize\sffamily,inner sep=0pt] at (4.550,-0.840)
  {190/190\\100\%};
\fill[fill={rgb,255:red,227;green,243;blue,153}] (5.500,-0.400) rectangle (7.400,-1.280);
\node[text=black,align=center,font=\scriptsize\sffamily,inner sep=0pt] at (6.450,-0.840)
  {189/210\\90\%};
\fill[fill={rgb,255:red,57;green,167;blue,88}] (7.400,-0.400) rectangle (9.300,-1.280);
\node[text=white,align=center,font=\scriptsize\sffamily,inner sep=0pt] at (8.350,-0.840)
  {180/180\\100\%};
\fill[fill={rgb,255:red,57;green,167;blue,88}] (9.300,-0.400) rectangle (11.200,-1.280);
\node[text=white,align=center,font=\scriptsize\sffamily,inner sep=0pt] at (10.250,-0.840)
  {110/110\\100\%};
\fill[fill={rgb,255:red,227;green,243;blue,153}] (11.200,-0.400) rectangle (13.100,-1.280);
\node[text=black,align=center,font=\scriptsize\sffamily,inner sep=0pt] at (12.150,-0.840)
  {171/190\\90\%};
\fill[fill={rgb,255:red,227;green,243;blue,153}] (13.100,-0.400) rectangle (15.000,-1.280);
\node[text=black,align=center,font=\scriptsize\sffamily,inner sep=0pt] at (14.050,-0.840)
  {126/140\\90\%};
\node[anchor=west,font=\small\sffamily,inner sep=3pt] at (15.120,-0.840)
  {$\!966$};
\draw[blue!60!black,line width=1.5pt] (3.600,-0.400) rectangle (5.500,-1.280);
\draw[blue!60!black,line width=1.5pt] (7.400,-0.400) rectangle (9.300,-1.280);
\draw[blue!60!black,line width=1.5pt] (9.300,-0.400) rectangle (11.200,-1.280);
\draw[blue!60!black,line width=1.5pt] (13.100,-0.400) rectangle (15.000,-1.280);
\fill[black!5] (0,-1.280) rectangle (3.600,-2.160);
\node[anchor=west,font=\small\sffamily,inner sep=2pt] at (0.10,-1.720) {TabICL};
\fill[fill={rgb,255:red,57;green,167;blue,88}] (3.600,-1.280) rectangle (5.500,-2.160);
\node[text=white,align=center,font=\scriptsize\sffamily,inner sep=0pt] at (4.550,-1.720)
  {190/190\\100\%};
\fill[fill={rgb,255:red,227;green,243;blue,153}] (5.500,-1.280) rectangle (7.400,-2.160);
\node[text=black,align=center,font=\scriptsize\sffamily,inner sep=0pt] at (6.450,-1.720)
  {189/210\\90\%};
\fill[fill={rgb,255:red,57;green,167;blue,88}] (7.400,-1.280) rectangle (9.300,-2.160);
\node[text=white,align=center,font=\scriptsize\sffamily,inner sep=0pt] at (8.350,-1.720)
  {180/180\\100\%};
\fill[fill={rgb,255:red,57;green,167;blue,88}] (9.300,-1.280) rectangle (11.200,-2.160);
\node[text=white,align=center,font=\scriptsize\sffamily,inner sep=0pt] at (10.250,-1.720)
  {110/110\\100\%};
\fill[fill={rgb,255:red,227;green,243;blue,153}] (11.200,-1.280) rectangle (13.100,-2.160);
\node[text=black,align=center,font=\scriptsize\sffamily,inner sep=0pt] at (12.150,-1.720)
  {171/190\\90\%};
\fill[fill={rgb,255:red,227;green,243;blue,153}] (13.100,-1.280) rectangle (15.000,-2.160);
\node[text=black,align=center,font=\scriptsize\sffamily,inner sep=0pt] at (14.050,-1.720)
  {126/140\\90\%};
\node[anchor=west,font=\small\sffamily,inner sep=3pt] at (15.120,-1.720)
  {$\!966$};
\draw[blue!60!black,line width=1.5pt] (3.600,-1.280) rectangle (5.500,-2.160);
\draw[blue!60!black,line width=1.5pt] (7.400,-1.280) rectangle (9.300,-2.160);
\draw[blue!60!black,line width=1.5pt] (9.300,-1.280) rectangle (11.200,-2.160);
\draw[blue!60!black,line width=1.5pt] (13.100,-1.280) rectangle (15.000,-2.160);
\fill[black!5] (0,-2.160) rectangle (3.600,-3.040);
\node[anchor=west,font=\small\sffamily,inner sep=2pt] at (0.10,-2.600) {MediTab};
\fill[fill={rgb,255:red,57;green,167;blue,88}] (3.600,-2.160) rectangle (5.500,-3.040);
\node[text=white,align=center,font=\scriptsize\sffamily,inner sep=0pt] at (4.550,-2.600)
  {190/190\\100\%};
\fill[fill={rgb,255:red,57;green,167;blue,88}] (5.500,-2.160) rectangle (7.400,-3.040);
\node[text=white,align=center,font=\scriptsize\sffamily,inner sep=0pt] at (6.450,-2.600)
  {210/210\\100\%};
\fill[fill={rgb,255:red,57;green,167;blue,88}] (7.400,-2.160) rectangle (9.300,-3.040);
\node[text=white,align=center,font=\scriptsize\sffamily,inner sep=0pt] at (8.350,-2.600)
  {180/180\\100\%};
\fill[fill={rgb,255:red,245;green,114;blue,69}] (9.300,-2.160) rectangle (11.200,-3.040);
\node[text=black,align=center,font=\scriptsize\sffamily,inner sep=0pt] at (10.250,-2.600)
  {85/110\\77\%};
\fill[fill={rgb,255:red,227;green,243;blue,153}] (11.200,-2.160) rectangle (13.100,-3.040);
\node[text=black,align=center,font=\scriptsize\sffamily,inner sep=0pt] at (12.150,-2.600)
  {171/190\\90\%};
\fill[fill={rgb,255:red,227;green,243;blue,153}] (13.100,-2.160) rectangle (15.000,-3.040);
\node[text=black,align=center,font=\scriptsize\sffamily,inner sep=0pt] at (14.050,-2.600)
  {126/140\\90\%};
\node[anchor=west,font=\small\sffamily,inner sep=3pt] at (15.120,-2.600)
  {$\!962$};
\draw[blue!60!black,line width=1.5pt] (3.600,-2.160) rectangle (5.500,-3.040);
\draw[blue!60!black,line width=1.5pt] (5.500,-2.160) rectangle (7.400,-3.040);
\draw[blue!60!black,line width=1.5pt] (7.400,-2.160) rectangle (9.300,-3.040);
\draw[blue!60!black,line width=1.5pt] (13.100,-2.160) rectangle (15.000,-3.040);
\fill[black!5] (0,-3.040) rectangle (3.600,-3.920);
\node[anchor=west,font=\small\sffamily,inner sep=2pt] at (0.10,-3.480) {TAPTAP};
\fill[fill={rgb,255:red,57;green,167;blue,88}] (3.600,-3.040) rectangle (5.500,-3.920);
\node[text=white,align=center,font=\scriptsize\sffamily,inner sep=0pt] at (4.550,-3.480)
  {190/190\\100\%};
\fill[fill={rgb,255:red,57;green,167;blue,88}] (5.500,-3.040) rectangle (7.400,-3.920);
\node[text=white,align=center,font=\scriptsize\sffamily,inner sep=0pt] at (6.450,-3.480)
  {210/210\\100\%};
\fill[fill={rgb,255:red,57;green,167;blue,88}] (7.400,-3.040) rectangle (9.300,-3.920);
\node[text=white,align=center,font=\scriptsize\sffamily,inner sep=0pt] at (8.350,-3.480)
  {180/180\\100\%};
\fill[fill={rgb,255:red,216;green,49;blue,40}] (9.300,-3.040) rectangle (11.200,-3.920);
\node[text=white,align=center,font=\scriptsize\sffamily,inner sep=0pt] at (10.250,-3.480)
  {81/110\\74\%};
\fill[fill={rgb,255:red,227;green,243;blue,153}] (11.200,-3.040) rectangle (13.100,-3.920);
\node[text=black,align=center,font=\scriptsize\sffamily,inner sep=0pt] at (12.150,-3.480)
  {171/190\\90\%};
\fill[fill={rgb,255:red,227;green,243;blue,153}] (13.100,-3.040) rectangle (15.000,-3.920);
\node[text=black,align=center,font=\scriptsize\sffamily,inner sep=0pt] at (14.050,-3.480)
  {126/140\\90\%};
\node[anchor=west,font=\small\sffamily,inner sep=3pt] at (15.120,-3.480)
  {$\!958$};
\draw[blue!60!black,line width=1.5pt] (3.600,-3.040) rectangle (5.500,-3.920);
\draw[blue!60!black,line width=1.5pt] (5.500,-3.040) rectangle (7.400,-3.920);
\draw[blue!60!black,line width=1.5pt] (7.400,-3.040) rectangle (9.300,-3.920);
\draw[blue!60!black,line width=1.5pt] (13.100,-3.040) rectangle (15.000,-3.920);
\fill[black!5] (0,-3.920) rectangle (3.600,-4.800);
\node[anchor=west,font=\small\sffamily,inner sep=2pt] at (0.10,-4.360) {VIME};
\fill[fill={rgb,255:red,102;green,189;blue,99}] (3.600,-3.920) rectangle (5.500,-4.800);
\node[text=black,align=center,font=\scriptsize\sffamily,inner sep=0pt] at (4.550,-4.360)
  {186/190\\98\%};
\fill[fill={rgb,255:red,227;green,243;blue,153}] (5.500,-3.920) rectangle (7.400,-4.800);
\node[text=black,align=center,font=\scriptsize\sffamily,inner sep=0pt] at (6.450,-4.360)
  {189/210\\90\%};
\fill[fill={rgb,255:red,57;green,167;blue,88}] (7.400,-3.920) rectangle (9.300,-4.800);
\node[text=white,align=center,font=\scriptsize\sffamily,inner sep=0pt] at (8.350,-4.360)
  {180/180\\100\%};
\fill[fill={rgb,255:red,132;green,202;blue,102}] (9.300,-3.920) rectangle (11.200,-4.800);
\node[text=black,align=center,font=\scriptsize\sffamily,inner sep=0pt] at (10.250,-4.360)
  {106/110\\96\%};
\fill[fill={rgb,255:red,227;green,243;blue,153}] (11.200,-3.920) rectangle (13.100,-4.800);
\node[text=black,align=center,font=\scriptsize\sffamily,inner sep=0pt] at (12.150,-4.360)
  {171/190\\90\%};
\fill[fill={rgb,255:red,227;green,243;blue,153}] (13.100,-3.920) rectangle (15.000,-4.800);
\node[text=black,align=center,font=\scriptsize\sffamily,inner sep=0pt] at (14.050,-4.360)
  {126/140\\90\%};
\node[anchor=west,font=\small\sffamily,inner sep=3pt] at (15.120,-4.360)
  {$\!958$};
\draw[blue!60!black,line width=1.5pt] (7.400,-3.920) rectangle (9.300,-4.800);
\draw[blue!60!black,line width=1.5pt] (13.100,-3.920) rectangle (15.000,-4.800);
\fill[black!5] (0,-4.800) rectangle (3.600,-5.680);
\node[anchor=west,font=\small\sffamily,inner sep=2pt] at (0.10,-5.240) {SAINT};
\fill[fill={rgb,255:red,102;green,189;blue,99}] (3.600,-4.800) rectangle (5.500,-5.680);
\node[text=black,align=center,font=\scriptsize\sffamily,inner sep=0pt] at (4.550,-5.240)
  {186/190\\98\%};
\fill[fill={rgb,255:red,227;green,243;blue,153}] (5.500,-4.800) rectangle (7.400,-5.680);
\node[text=black,align=center,font=\scriptsize\sffamily,inner sep=0pt] at (6.450,-5.240)
  {189/210\\90\%};
\fill[fill={rgb,255:red,105;green,190;blue,99}] (7.400,-4.800) rectangle (9.300,-5.680);
\node[text=black,align=center,font=\scriptsize\sffamily,inner sep=0pt] at (8.350,-5.240)
  {176/180\\98\%};
\fill[fill={rgb,255:red,216;green,49;blue,40}] (9.300,-4.800) rectangle (11.200,-5.680);
\node[text=white,align=center,font=\scriptsize\sffamily,inner sep=0pt] at (10.250,-5.240)
  {81/110\\74\%};
\fill[fill={rgb,255:red,57;green,167;blue,88}] (11.200,-4.800) rectangle (13.100,-5.680);
\node[text=white,align=center,font=\scriptsize\sffamily,inner sep=0pt] at (12.150,-5.240)
  {190/190\\100\%};
\fill[fill={rgb,255:red,227;green,243;blue,153}] (13.100,-4.800) rectangle (15.000,-5.680);
\node[text=black,align=center,font=\scriptsize\sffamily,inner sep=0pt] at (14.050,-5.240)
  {126/140\\90\%};
\node[anchor=west,font=\small\sffamily,inner sep=3pt] at (15.120,-5.240)
  {$\!948$};
\draw[blue!60!black,line width=1.5pt] (11.200,-4.800) rectangle (13.100,-5.680);
\draw[blue!60!black,line width=1.5pt] (13.100,-4.800) rectangle (15.000,-5.680);
\fill[black!5] (0,-5.680) rectangle (3.600,-6.560);
\node[anchor=west,font=\small\sffamily,inner sep=2pt] at (0.10,-6.120) {TabDPT};
\fill[fill={rgb,255:red,142;green,207;blue,103}] (3.600,-5.680) rectangle (5.500,-6.560);
\node[text=black,align=center,font=\scriptsize\sffamily,inner sep=0pt] at (4.550,-6.120)
  {182/190\\96\%};
\fill[fill={rgb,255:red,227;green,243;blue,153}] (5.500,-5.680) rectangle (7.400,-6.560);
\node[text=black,align=center,font=\scriptsize\sffamily,inner sep=0pt] at (6.450,-6.120)
  {189/210\\90\%};
\fill[fill={rgb,255:red,147;green,209;blue,104}] (7.400,-5.680) rectangle (9.300,-6.560);
\node[text=black,align=center,font=\scriptsize\sffamily,inner sep=0pt] at (8.350,-6.120)
  {172/180\\96\%};
\fill[fill={rgb,255:red,132;green,202;blue,102}] (9.300,-5.680) rectangle (11.200,-6.560);
\node[text=black,align=center,font=\scriptsize\sffamily,inner sep=0pt] at (10.250,-6.120)
  {106/110\\96\%};
\fill[fill={rgb,255:red,227;green,243;blue,153}] (11.200,-5.680) rectangle (13.100,-6.560);
\node[text=black,align=center,font=\scriptsize\sffamily,inner sep=0pt] at (12.150,-6.120)
  {171/190\\90\%};
\fill[fill={rgb,255:red,227;green,243;blue,153}] (13.100,-5.680) rectangle (15.000,-6.560);
\node[text=black,align=center,font=\scriptsize\sffamily,inner sep=0pt] at (14.050,-6.120)
  {126/140\\90\%};
\node[anchor=west,font=\small\sffamily,inner sep=3pt] at (15.120,-6.120)
  {$\!946$};
\draw[blue!60!black,line width=1.5pt] (13.100,-5.680) rectangle (15.000,-6.560);
\fill[black!5] (0,-6.560) rectangle (3.600,-7.440);
\node[anchor=west,font=\small\sffamily,inner sep=2pt] at (0.10,-7.000) {XTab};
\fill[fill={rgb,255:red,57;green,167;blue,88}] (3.600,-6.560) rectangle (5.500,-7.440);
\node[text=white,align=center,font=\scriptsize\sffamily,inner sep=0pt] at (4.550,-7.000)
  {190/190\\100\%};
\fill[fill={rgb,255:red,227;green,243;blue,153}] (5.500,-6.560) rectangle (7.400,-7.440);
\node[text=black,align=center,font=\scriptsize\sffamily,inner sep=0pt] at (6.450,-7.000)
  {189/210\\90\%};
\fill[fill={rgb,255:red,227;green,243;blue,153}] (7.400,-6.560) rectangle (9.300,-7.440);
\node[text=black,align=center,font=\scriptsize\sffamily,inner sep=0pt] at (8.350,-7.000)
  {162/180\\90\%};
\fill[fill={rgb,255:red,227;green,243;blue,153}] (9.300,-6.560) rectangle (11.200,-7.440);
\node[text=black,align=center,font=\scriptsize\sffamily,inner sep=0pt] at (10.250,-7.000)
  {99/110\\90\%};
\fill[fill={rgb,255:red,227;green,243;blue,153}] (11.200,-6.560) rectangle (13.100,-7.440);
\node[text=black,align=center,font=\scriptsize\sffamily,inner sep=0pt] at (12.150,-7.000)
  {171/190\\90\%};
\fill[fill={rgb,255:red,227;green,243;blue,153}] (13.100,-6.560) rectangle (15.000,-7.440);
\node[text=black,align=center,font=\scriptsize\sffamily,inner sep=0pt] at (14.050,-7.000)
  {126/140\\90\%};
\node[anchor=west,font=\small\sffamily,inner sep=3pt] at (15.120,-7.000)
  {$\!937$};
\draw[blue!60!black,line width=1.5pt] (3.600,-6.560) rectangle (5.500,-7.440);
\draw[blue!60!black,line width=1.5pt] (13.100,-6.560) rectangle (15.000,-7.440);

\fill[fill={rgb,255:red,171;green,6;blue,38}] (16.600,-7.440) rectangle (16.980,-7.254);
\fill[fill={rgb,255:red,183;green,17;blue,38}] (16.600,-7.254) rectangle (16.980,-7.068);
\fill[fill={rgb,255:red,196;green,30;blue,39}] (16.600,-7.068) rectangle (16.980,-6.882);
\fill[fill={rgb,255:red,208;green,41;blue,39}] (16.600,-6.882) rectangle (16.980,-6.696);
\fill[fill={rgb,255:red,218;green,54;blue,42}] (16.600,-6.696) rectangle (16.980,-6.510);
\fill[fill={rgb,255:red,226;green,71;blue,49}] (16.600,-6.510) rectangle (16.980,-6.324);
\fill[fill={rgb,255:red,233;green,85;blue,56}] (16.600,-6.324) rectangle (16.980,-6.138);
\fill[fill={rgb,255:red,241;green,102;blue,64}] (16.600,-6.138) rectangle (16.980,-5.952);
\fill[fill={rgb,255:red,245;green,117;blue,71}] (16.600,-5.952) rectangle (16.980,-5.766);
\fill[fill={rgb,255:red,247;green,132;blue,78}] (16.600,-5.766) rectangle (16.980,-5.580);
\fill[fill={rgb,255:red,250;green,150;blue,86}] (16.600,-5.580) rectangle (16.980,-5.394);
\fill[fill={rgb,255:red,252;green,165;blue,93}] (16.600,-5.394) rectangle (16.980,-5.208);
\fill[fill={rgb,255:red,253;green,181;blue,103}] (16.600,-5.208) rectangle (16.980,-5.022);
\fill[fill={rgb,255:red,253;green,193;blue,113}] (16.600,-5.022) rectangle (16.980,-4.836);
\fill[fill={rgb,255:red,254;green,204;blue,123}] (16.600,-4.836) rectangle (16.980,-4.650);
\fill[fill={rgb,255:red,254;green,218;blue,134}] (16.600,-4.650) rectangle (16.980,-4.464);
\fill[fill={rgb,255:red,254;green,228;blue,145}] (16.600,-4.464) rectangle (16.980,-4.278);
\fill[fill={rgb,255:red,254;green,236;blue,159}] (16.600,-4.278) rectangle (16.980,-4.092);
\fill[fill={rgb,255:red,255;green,243;blue,172}] (16.600,-4.092) rectangle (16.980,-3.906);
\fill[fill={rgb,255:red,255;green,251;blue,184}] (16.600,-3.906) rectangle (16.980,-3.720);
\fill[fill={rgb,255:red,250;green,253;blue,184}] (16.600,-3.720) rectangle (16.980,-3.534);
\fill[fill={rgb,255:red,241;green,249;blue,172}] (16.600,-3.534) rectangle (16.980,-3.348);
\fill[fill={rgb,255:red,230;green,245;blue,157}] (16.600,-3.348) rectangle (16.980,-3.162);
\fill[fill={rgb,255:red,221;green,241;blue,145}] (16.600,-3.162) rectangle (16.980,-2.976);
\fill[fill={rgb,255:red,211;green,236;blue,135}] (16.600,-2.976) rectangle (16.980,-2.790);
\fill[fill={rgb,255:red,197;green,230;blue,126}] (16.600,-2.790) rectangle (16.980,-2.604);
\fill[fill={rgb,255:red,185;green,225;blue,118}] (16.600,-2.604) rectangle (16.980,-2.418);
\fill[fill={rgb,255:red,171;green,219;blue,109}] (16.600,-2.418) rectangle (16.980,-2.232);
\fill[fill={rgb,255:red,157;green,213;blue,105}] (16.600,-2.232) rectangle (16.980,-2.046);
\fill[fill={rgb,255:red,142;green,207;blue,103}] (16.600,-2.046) rectangle (16.980,-1.860);
\fill[fill={rgb,255:red,125;green,199;blue,101}] (16.600,-1.860) rectangle (16.980,-1.674);
\fill[fill={rgb,255:red,110;green,192;blue,100}] (16.600,-1.674) rectangle (16.980,-1.488);
\fill[fill={rgb,255:red,90;green,183;blue,96}] (16.600,-1.488) rectangle (16.980,-1.302);
\fill[fill={rgb,255:red,72;green,174;blue,92}] (16.600,-1.302) rectangle (16.980,-1.116);
\fill[fill={rgb,255:red,54;green,166;blue,87}] (16.600,-1.116) rectangle (16.980,-0.930);
\fill[fill={rgb,255:red,33;green,156;blue,82}] (16.600,-0.930) rectangle (16.980,-0.744);
\fill[fill={rgb,255:red,22;green,145;blue,77}] (16.600,-0.744) rectangle (16.980,-0.558);
\fill[fill={rgb,255:red,15;green,132;blue,70}] (16.600,-0.558) rectangle (16.980,-0.372);
\fill[fill={rgb,255:red,9;green,121;blue,64}] (16.600,-0.372) rectangle (16.980,-0.186);
\fill[fill={rgb,255:red,3;green,110;blue,58}] (16.600,-0.186) rectangle (16.980,-0.000);
\draw[black,line width=0.5pt] (16.600,-7.440) rectangle (16.980,0.000);
\draw[black,line width=0.5pt] (16.980,-7.440)--(17.130,-7.440);
\node[anchor=west,font=\tiny\sffamily] at (17.160,-7.440) {70\%};
\draw[black,line width=0.5pt] (16.980,-6.377)--(17.130,-6.377);
\node[anchor=west,font=\tiny\sffamily] at (17.160,-6.377) {75\%};
\draw[black,line width=0.5pt] (16.980,-5.314)--(17.130,-5.314);
\node[anchor=west,font=\tiny\sffamily] at (17.160,-5.314) {80\%};
\draw[black,line width=0.5pt] (16.980,-4.251)--(17.130,-4.251);
\node[anchor=west,font=\tiny\sffamily] at (17.160,-4.251) {85\%};
\draw[black,line width=0.5pt] (16.980,-3.189)--(17.130,-3.189);
\node[anchor=west,font=\tiny\sffamily] at (17.160,-3.189) {90\%};
\draw[black,line width=0.5pt] (16.980,-2.126)--(17.130,-2.126);
\node[anchor=west,font=\tiny\sffamily] at (17.160,-2.126) {95\%};
\draw[black,line width=0.5pt] (16.980,-1.063)--(17.130,-1.063);
\node[anchor=west,font=\tiny\sffamily] at (17.160,-1.063) {100\%};
\node[rotate=90,anchor=south,font=\scriptsize\sffamily] at (17.880,-3.720) {Score / max possible};

\end{tikzpicture}
\caption{Metric-layer performance profile under Iron Deficiency use-case weights, computed from the independent-literature-based property audit (Section~\ref{sec:audit-note-v2}). Each cell shows the raw metric score relative to the use-case-adjusted maximum and the corresponding percentage. Blue borders mark the highest-scoring model in each metric column. The eight models shown are the eight with the strongest combined rank across both use cases (Section~\ref{sec:hf-usecase}); no model is pre-selected to illustrate a trade-off. $\mathcal{R}$ is the overall weighted ranking score (max\,=\,1020).}
\label{fig:id_heatmap}
\end{figure}
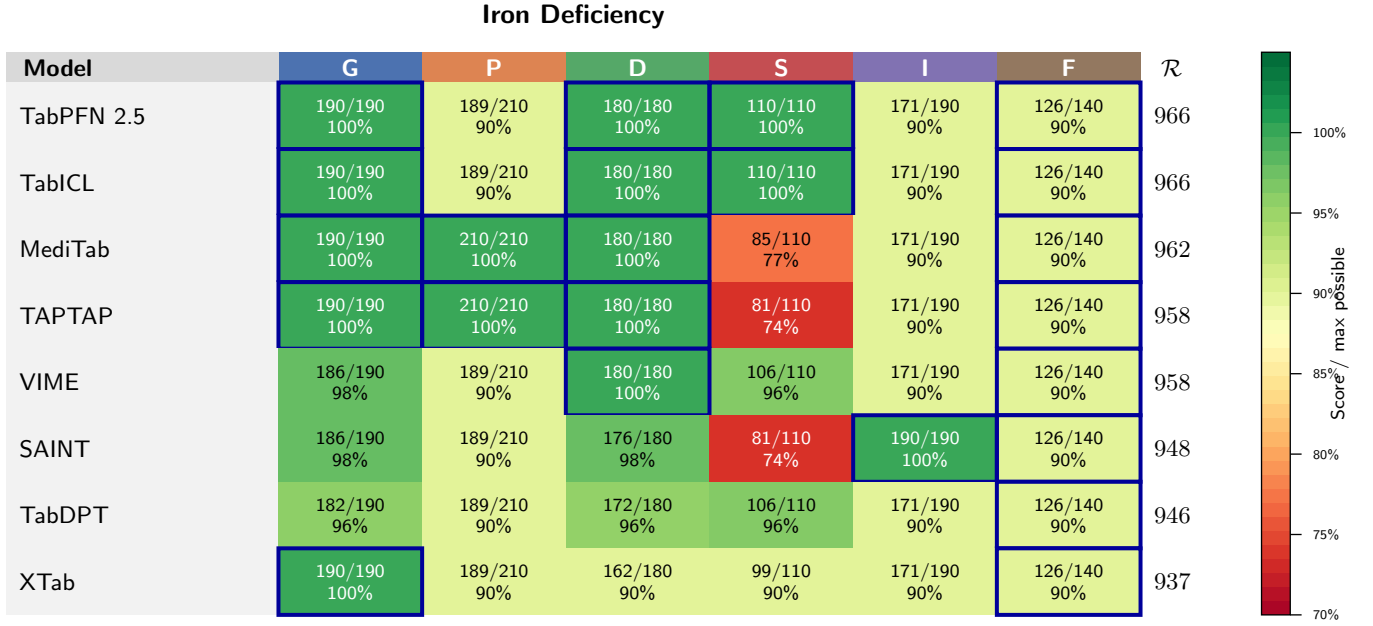

Applying these sub-metric weights to the scores in Figure~\ref{fig:tfm_scores} produces the second and third layers of the framework simultaneously. Figure~\ref{fig:id_heatmap} shows the resulting metric-layer profile for each model: each cell reports a model's weighted score on one metric as a fraction of the use-case maximum, making it immediately visible where a model gains or loses ground under these clinical priorities. In Table~\ref{tab:id_ranking}, the six columns $\mathcal{G}$, $\mathcal{P}$, $\mathcal{D}$, $\mathcal{S}$, $\mathcal{I}$, and $\mathcal{F}$ are the same metric-layer scores in numerical form. The final column $\mathcal{R}(\tau_i)$ is the overall ranking, the sum across all six metric scores, which integrates the full sub-metric profile into a single number for direct model comparison.

\begin{table}[t]
\centering
\small
\caption{Model ranking for the iron deficiency use case, computed from the audited property assignments (Section~\ref{sec:audit-note}). Scores represent the weighted sum $\mathcal{R}(\tau_i)$ across all sub-metrics under the iron deficiency sub-metric weights. The maximum possible score is 1020. The eight models shown are the eight with the strongest combined rank across both use cases (Section~\ref{sec:hf-usecase}); no model is pre-selected to illustrate a trade-off.}
\label{tab:id_ranking}
\begin{tabular}{|l|c|c|c|c|c|c|c|}
\hline
\textbf{Model} & $\mathcal{G}$ & $\mathcal{P}$ & $\mathcal{D}$ & $\mathcal{S}$ & $\mathcal{I}$ & $\mathcal{F}$ & $\mathcal{R}(\tau_i)$ \\
\hline
Max possible    & 190 & 210 & 180 & 110 & 190 & 140 & 1020 \\ \hline
TabPFN~2.5      & 190 & 189 & 180 & 110 & 171 & 126 & \textbf{966} \\
TabICL          & 190 & 189 & 180 & 110 & 171 & 126 & \textbf{966} \\
MediTab         & 190 & 210 & 180 &  85 & 171 & 126 &  962 \\
TAPTAP          & 190 & 210 & 180 &  81 & 171 & 126 &  958 \\
VIME            & 186 & 189 & 180 & 106 & 171 & 126 &  958 \\
SAINT           & 186 & 189 & 176 &  81 & 190 & 126 &  948 \\
TabDPT          & 182 & 189 & 172 & 106 & 171 & 126 &  946 \\
XTab            & 190 & 189 & 162 &  99 & 171 & 126 &  937 \\
\hline
\end{tabular}
\end{table}

TabPFN~2.5 and TabICL tie for first place at 966, an exact tie across every one of the 25 sub-metrics that compose the ranking, not merely a coincidence in the weighted total. This tie is itself a finding of the second, independent-literature-based audit pass: TabPFN~2.5's Handle Noisy Inputs property, previously scored absent on the strength of its own technical report alone, is scored present following an independent finding from TabArena (Erickson et al., NeurIPS 2025 D\&B), a large, realistic benchmark suite reporting that ``TabPFN-2.5 is the leading method... substantially outperforming tuned tree-based models,'' consistent with the robust-scaling and soft-clipping mechanisms its own report describes. This closes the single sub-metric, Data Noise Robustness, on which the two models previously differed (9 versus 10), so the two are reported here as co-leaders rather than one ranked ahead of the other.

MediTab places third at 962, again the only model in the top eight to reach the privacy ceiling (210/210), but its resource-utilization, computational-scalability, and training-time-efficiency sub-metrics remain the weakest in the table (85/110). TAPTAP and VIME tie for fourth at 958. TAPTAP reaches the same privacy ceiling as MediTab and, like MediTab, gains a schema-robustness and data-noise-robustness point from a Handle Missing Data property newly scored present on the strength of its own paper's imputation use case, though this specific flip is flagged low/moderate confidence pending independent confirmation. TAPTAP's privacy ceiling itself, unlike MediTab's, is not free of dispute: an independent study of memorization in GPT-2-derived tabular generators found measurable numeric-string leakage in the plain GPT-2 backbone that TAPTAP continues pretraining from -- an indirect but unresolved challenge to the ``present'' anonymization reading retained here under this audit's conflict-resolution rule, so TAPTAP's place in this table should be read as provisional rather than settled (Section~\ref{sec:audit-note}). VIME reaches the same total through its lightweight self-supervised architecture, matching the leaders on generalizability and data-efficient learning without a comparable privacy ceiling.

SAINT, at 948, is the only model in this table to reach the interpretability ceiling (190/190), reflecting its paper-documented attention-based interpretability claim, but an independent benchmark (McElfresh et al., ``TabZilla,'' NeurIPS 2023 D\&B) directly measured SAINT's training time as the slowest of 19 tested algorithms, confirming rather than merely leaving ambiguous the absence of its resource-utilization and embedding-generation properties and producing the lowest scalability score in the top eight (81/110). TabDPT (946) and XTab (937) close out the table; neither model's scores changed under this audit, since no independent literature addressing either model specifically was found in any of the domain files reviewed, and their presence in the top eight is a consequence of TabPFN~v2, TabPFN, and TabNet falling out of contention rather than new evidence in either model's favor (Section~\ref{sec:audit-note}).

The overall ranking $\mathcal{R}(\tau_i)$ is the primary basis for model selection. For readers who want additional diagnostic depth, Table~\ref{tab:id_supermetrics} shows the four supplementary super-metric scores for all eight models under the iron deficiency sub-metric weights. These scores do not change the recommendation but reveal how each model's strengths are distributed across compound cross-domain concerns.

\begin{table}[t]
\centering
\small
\caption{Super-metric scores for the iron deficiency use case. TLC = Transfer Learning Capacity ($\mathcal{TLC}$, max 340), MR = Model Resilience ($\mathcal{MR}$, max 170), EPT = Efficiency-Privacy Trade-off ($\mathcal{EPT}$, max 500), CT = Clinical Trustworthiness ($\mathcal{CT}$, max 330). Scores are weighted sums using the iron deficiency sub-metric weights.}
\label{tab:id_supermetrics}
\begin{tabular}{|l|c|c|c|c|}
\hline
\textbf{Model} & \textbf{TLC} (340) & \textbf{MR} (170) & \textbf{EPT} (500) & \textbf{CT} (330) \\
\hline
TabPFN~2.5      & 333 & 166 & 479 & 297 \\
TabICL          & 333 & 166 & 479 & 297 \\
MediTab         & 340 & 170 & 475 & 297 \\
TAPTAP          & 340 & 170 & 471 & 297 \\
VIME            & 329 & 162 & 475 & 297 \\
SAINT           & 325 & 158 & 446 & 316 \\
TabDPT          & 317 & 150 & 467 & 297 \\
XTab            & 325 & 162 & 450 & 297 \\
\hline
\end{tabular}
\end{table}

On Transfer Learning Capacity, MediTab and TAPTAP now lead at the ceiling (340/340), each combining maximal generalizability with a perfect privacy profile and the same Handle Missing Data-driven noise-robustness point. TabPFN~2.5 and TabICL follow at 333, held one point short of the ceiling by their privacy sub-metrics despite maximal generalizability and data-efficiency components. VIME reaches 329. SAINT and XTab tie at 325, each held back by a schema-robustness or data-noise-robustness score of 9 within the TLC composition. TabDPT trails at 317, the only model in the table with both its schema-robustness and data-noise-robustness scores at 8, neither of which this audit found independent evidence to revise.

Model Resilience follows a similar pattern: MediTab and TAPTAP lead at the ceiling (170/170), TabPFN~2.5 and TabICL follow at 166, VIME and XTab tie at 162, SAINT trails at 158, and TabDPT is lowest at 150, reflecting its unrevised schema-robustness and data-noise-robustness scores.

The Efficiency-Privacy Trade-off widens the picture to include all scalability sub-metrics alongside privacy and data-efficiency. TabPFN~2.5 and TabICL lead at 479/500, benefiting from their maximal scalability scores. MediTab and VIME follow at 475, TAPTAP at 471, TabDPT at 467, XTab at 450, and SAINT trails at 446, the lowest in the table, penalized by the resource-utilization and computational-scalability weakness independently documented in Section~\ref{sec:id-usecase}.

Clinical Trustworthiness splits the table along a single line: SAINT alone reaches 316/330, the only model in the top eight with a documented, paper-positioned interpretability mechanism (attention-based) confirmed as a ceiling-level claim by its own paper. The remaining seven models score 297/330, each held to the interpretability sub-metric floor of 9 rather than 10 across FIC, MTS, EC, and LI.

For an iron deficiency screening program, \textbf{TabPFN~2.5 and TabICL are the primary joint recommendation}: the two models are tied for the overall ranking at 966/1020 with an identical 25-sub-metric profile, so this framework's output gives no principled basis for preferring one over the other. MediTab is the strongest alternative at 962, offering a privacy profile neither co-leader matches at the cost of the weakest scalability score in the table. SAINT is the closest match on Clinical Trustworthiness specifically, reaching the interpretability ceiling that the co-leaders do not, again at the cost of the weakest scalability score in the top eight.

\subsection{Heart Failure Use Case}
\label{sec:hf-usecase}

Heart failure is a clinical syndrome affecting tens of millions of patients worldwide, characterized by the heart's inability to meet the body's circulatory demands, and it carries a burden of repeated hospitalization, declining functional capacity, and high mortality that places it among the most resource-intensive conditions in any healthcare system. The challenge for clinical AI is not simply to predict who will be hospitalized, but to do so early enough, accurately enough, and with enough transparency that clinicians can act on the prediction with confidence. Recent work has shown how demanding this combination of requirements is in practice. Gao and colleagues developed a multimodal deep learning model combining structured clinical tabular data with unstructured admission notes from EHR records, demonstrating that integrating both data types substantially improved the prognostic accuracy for in-hospital mortality and severe decompensation compared to either modality alone \cite{gao2024improving}. Najarian and colleagues used interpretable logistic tensor regression on sequential EHR records to predict eligibility for advanced therapies including transplantation and mechanical circulatory support, achieving an AUC of 0.903 and identifying chronic kidney disease, pulmonary heart disease, and mitral regurgitation as the most influential clinical codes, with findings validated directly by heart failure cardiologists \cite{najarian2024enhancing}. Fine and colleagues applied a gradient-boosted model (CatBoost) to administrative health data from over 50,000 patients to predict 30-day and one-year outcomes after heart failure emergency department visits and hospital admissions, demonstrating that even administrative data, without detailed clinical variables, contains sufficient signal for risk stratification \cite{fine2024machine}. Moreno-Sánchez applied explainable AI methods including SHAP values to a heart failure survival dataset, showing that post-hoc explanation tools can meaningfully improve the interpretability and clinical acceptance of survival prediction models \cite{moreno2023improvement}. On the fairness side, Sufian and colleagues examined algorithmic bias in cardiovascular AI systems and demonstrated that standard fairness metrics such as disparate impact and equal opportunity difference could be substantially improved through fairness-aware training, with practical gains for both male and female patient groups \cite{sufian2024mitigating}. These studies collectively illustrate the landscape in which a tabular foundation model for heart failure would be deployed: large and heterogeneous EHR datasets, high interpretability expectations from cardiologists, strong privacy and regulatory requirements, and documented demographic disparities that make fairness a first-order concern rather than an afterthought.

Applying \system{} to this use case again begins at the metric level, and the priorities shift in ways that reflect the different clinical context. Privacy Preservation is again the highest-priority metric, but it is weighted more heavily than in the iron deficiency case because multi-site heart failure studies involve detailed longitudinal cardiac records that are more re-identifiable than routine blood count panels. Generalizability is the second priority, because multi-site deployment across institutions with heterogeneous EHR schemas is the standard setting for heart failure AI research. Interpretability is elevated equally: cardiologists managing patients with advanced heart failure will not act on a prediction they cannot understand. Scalability rises substantially compared to the iron deficiency case, because heart failure datasets are genuinely large and complex, spanning medications, diagnoses, biomarkers, echocardiographic measurements, and vital signs. Fairness is elevated to its highest level in this framework application, with all three sub-metrics at maximum weight, reflecting the well-documented disparities in heart failure outcomes across age, sex, and racial groups. Data-Efficient Learning, while still important, receives the lowest combined priority, because the relative data abundance in large hospital cardiology programs reduces the urgency of few-shot and augmentation capabilities. These metric-level priorities then determine the sub-metric weights within each dimension.

Under the Generalizability metric, Intra-Domain Transferability (IDT) receives a weight of 5, because heart failure datasets from different institutions reflect different coding practices, patient populations, and included variables, and a model that cannot generalize across these differences will require separate retraining for each deployment site. Cross-Domain Transferability (CDT) receives a weight of 3, as the cardiovascular domain shares relevant features with related conditions, making some cross-domain generalization practically useful. Schema Robustness (SR) receives the highest weight of 5, because EHR schemas change when hospital systems are upgraded and vary substantially across institutions, and a model that fails when columns are renamed or restructured is not viable for multi-site deployment. Feature Adaptation Capability (FAC) receives a weight of 4, reflecting the variability in clinical feature sets across heart failure cohorts depending on whether echocardiographic, biomarker, or administrative data is available. Zero-Shot Inference Adaptability (ZIA) receives a weight of 3, as in the iron deficiency case, since some fine-tuning is typically feasible.

Under the Privacy Preservation metric, all sub-metrics are elevated relative to the iron deficiency case. Membership Inference Resistance (MIR) receives a weight of 5 rather than 4, reflecting the particularly sensitive nature of detailed cardiac histories and the institutional emphasis on preventing re-identification in multi-site studies. Data Anonymization Robustness (DAR) also receives a weight of 5, since heart failure datasets are routinely de-identified before any sharing or collaborative use. Differential Privacy Compliance (DPC), Data Reconstruction Attack Resistance (DRAR), and Third-Party Data Handling Compliance (TPDHC) each receive a weight of 4, reflecting the multi-site collaborative environment in which heart failure prediction models are typically developed and validated.

Under the Data-Efficient Learning metric, the weights differ meaningfully from the iron deficiency case. Data Noise Robustness (DNR) receives the highest weight of 5, because EHR data for heart failure is well known for its missing values, inconsistent coding, and measurement noise, and a model that cannot tolerate these imperfections will struggle in real-world deployment. Performance with Reduced Data (PRD) receives a weight of 4, since many deployment sites, particularly community hospitals and regional centers, have moderate-sized rather than large datasets. Few-Shot Learning Capability (FLC) receives a weight of 3, as the relatively higher data availability in large hospital systems reduces its priority compared to the iron deficiency case. Data Augmentation Sensitivity (DAS) receives a weight of 3.

Under the Scalability metric, the weights are generally higher than in the iron deficiency case, reflecting the larger scale of heart failure datasets and the more complex feature spaces involved. Feature Scalability (FS) receives the highest weight of 5, because EHR-derived feature sets for heart failure can be highly dimensional, spanning medications, diagnoses, laboratory values, and vital signs, and the model must not degrade as feature complexity increases. Resource Utilization Efficiency (RUE) and Computational Scalability (CS) both receive a weight of 4, as clinical decision support systems must operate within hospital IT infrastructure constraints and heart failure surveillance involves large patient populations. Training Time Efficiency (TTE) receives a weight of 3.

Under the Interpretability metric, requirements are at least as stringent as in the iron deficiency case. Feature Importance Clarity (FIC), Explanation Consistency (EC), and Local Interpretability (LI) each receive a weight of 5, reflecting that clinicians managing heart failure patients require clear, consistent, and patient-specific explanations before they will incorporate model predictions into treatment decisions. Model Transparency Score (MTS) receives a weight of 4, as in the iron deficiency case.

Under the Fairness and Bias Mitigation metric, all three sub-metrics receive a weight of 5, which is higher overall than in the iron deficiency case. Heart failure outcomes exhibit well-documented disparities across age, sex, and racial groups, and the risk that a model trained on historically biased EHR data will amplify those disparities is a first-order concern. Predictive Parity Bias Score (PPBS), Bias Amplification Score (BAS), and Intersectional Fairness Score (IFS) all receive maximum weighting to reflect this priority.

\begin{figure}[t]
\centering
\begin{tikzpicture}[x=1cm, y=1cm, font=\sffamily]

\node[font=\normalsize\bfseries\sffamily, anchor=south]
  at ({7.500}, 0.20) {Heart Failure};

\fill[mG] (3.600,0) rectangle (5.500,-0.400);
\node[text=white,font=\small\bfseries\sffamily] at (4.550,-0.200) {G};
\fill[mP] (5.500,0) rectangle (7.400,-0.400);
\node[text=white,font=\small\bfseries\sffamily] at (6.450,-0.200) {P};
\fill[mD] (7.400,0) rectangle (9.300,-0.400);
\node[text=white,font=\small\bfseries\sffamily] at (8.350,-0.200) {D};
\fill[mS] (9.300,0) rectangle (11.200,-0.400);
\node[text=white,font=\small\bfseries\sffamily] at (10.250,-0.200) {S};
\fill[mI] (11.200,0) rectangle (13.100,-0.400);
\node[text=white,font=\small\bfseries\sffamily] at (12.150,-0.200) {I};
\fill[mF] (13.100,0) rectangle (15.000,-0.400);
\node[text=white,font=\small\bfseries\sffamily] at (14.050,-0.200) {F};
\fill[black!15] (0,0) rectangle (3.600,-0.400);
\node[anchor=west,font=\small\bfseries\sffamily] at (0.10,-0.200) {Model};
\node[anchor=west,font=\small\bfseries\sffamily] at (15.150,-0.200) {$\mathcal{R}$};

\fill[black!5] (0,-0.400) rectangle (3.600,-1.280);
\node[anchor=west,font=\small\sffamily,inner sep=2pt] at (0.10,-0.840) {TabPFN~2.5};
\fill[fill={rgb,255:red,57;green,167;blue,88}] (3.600,-0.400) rectangle (5.500,-1.280);
\node[text=white,align=center,font=\scriptsize\sffamily,inner sep=0pt] at (4.550,-0.840)
  {200/200\\100\%};
\fill[fill={rgb,255:red,227;green,243;blue,153}] (5.500,-0.400) rectangle (7.400,-1.280);
\node[text=black,align=center,font=\scriptsize\sffamily,inner sep=0pt] at (6.450,-0.840)
  {198/220\\90\%};
\fill[fill={rgb,255:red,57;green,167;blue,88}] (7.400,-0.400) rectangle (9.300,-1.280);
\node[text=white,align=center,font=\scriptsize\sffamily,inner sep=0pt] at (8.350,-0.840)
  {150/150\\100\%};
\fill[fill={rgb,255:red,57;green,167;blue,88}] (9.300,-0.400) rectangle (11.200,-1.280);
\node[text=white,align=center,font=\scriptsize\sffamily,inner sep=0pt] at (10.250,-0.840)
  {160/160\\100\%};
\fill[fill={rgb,255:red,227;green,243;blue,153}] (11.200,-0.400) rectangle (13.100,-1.280);
\node[text=black,align=center,font=\scriptsize\sffamily,inner sep=0pt] at (12.150,-0.840)
  {171/190\\90\%};
\fill[fill={rgb,255:red,227;green,243;blue,153}] (13.100,-0.400) rectangle (15.000,-1.280);
\node[text=black,align=center,font=\scriptsize\sffamily,inner sep=0pt] at (14.050,-0.840)
  {135/150\\90\%};
\node[anchor=west,font=\small\sffamily,inner sep=3pt] at (15.120,-0.840)
  {$\!1014$};
\draw[blue!60!black,line width=1.5pt] (3.600,-0.400) rectangle (5.500,-1.280);
\draw[blue!60!black,line width=1.5pt] (7.400,-0.400) rectangle (9.300,-1.280);
\draw[blue!60!black,line width=1.5pt] (9.300,-0.400) rectangle (11.200,-1.280);
\draw[blue!60!black,line width=1.5pt] (13.100,-0.400) rectangle (15.000,-1.280);
\fill[black!5] (0,-1.280) rectangle (3.600,-2.160);
\node[anchor=west,font=\small\sffamily,inner sep=2pt] at (0.10,-1.720) {TabICL};
\fill[fill={rgb,255:red,57;green,167;blue,88}] (3.600,-1.280) rectangle (5.500,-2.160);
\node[text=white,align=center,font=\scriptsize\sffamily,inner sep=0pt] at (4.550,-1.720)
  {200/200\\100\%};
\fill[fill={rgb,255:red,227;green,243;blue,153}] (5.500,-1.280) rectangle (7.400,-2.160);
\node[text=black,align=center,font=\scriptsize\sffamily,inner sep=0pt] at (6.450,-1.720)
  {198/220\\90\%};
\fill[fill={rgb,255:red,57;green,167;blue,88}] (7.400,-1.280) rectangle (9.300,-2.160);
\node[text=white,align=center,font=\scriptsize\sffamily,inner sep=0pt] at (8.350,-1.720)
  {150/150\\100\%};
\fill[fill={rgb,255:red,57;green,167;blue,88}] (9.300,-1.280) rectangle (11.200,-2.160);
\node[text=white,align=center,font=\scriptsize\sffamily,inner sep=0pt] at (10.250,-1.720)
  {160/160\\100\%};
\fill[fill={rgb,255:red,227;green,243;blue,153}] (11.200,-1.280) rectangle (13.100,-2.160);
\node[text=black,align=center,font=\scriptsize\sffamily,inner sep=0pt] at (12.150,-1.720)
  {171/190\\90\%};
\fill[fill={rgb,255:red,227;green,243;blue,153}] (13.100,-1.280) rectangle (15.000,-2.160);
\node[text=black,align=center,font=\scriptsize\sffamily,inner sep=0pt] at (14.050,-1.720)
  {135/150\\90\%};
\node[anchor=west,font=\small\sffamily,inner sep=3pt] at (15.120,-1.720)
  {$\!1014$};
\draw[blue!60!black,line width=1.5pt] (3.600,-1.280) rectangle (5.500,-2.160);
\draw[blue!60!black,line width=1.5pt] (7.400,-1.280) rectangle (9.300,-2.160);
\draw[blue!60!black,line width=1.5pt] (9.300,-1.280) rectangle (11.200,-2.160);
\draw[blue!60!black,line width=1.5pt] (13.100,-1.280) rectangle (15.000,-2.160);
\fill[black!5] (0,-2.160) rectangle (3.600,-3.040);
\node[anchor=west,font=\small\sffamily,inner sep=2pt] at (0.10,-2.600) {VIME};
\fill[fill={rgb,255:red,110;green,192;blue,100}] (3.600,-2.160) rectangle (5.500,-3.040);
\node[text=black,align=center,font=\scriptsize\sffamily,inner sep=0pt] at (4.550,-2.600)
  {195/200\\98\%};
\fill[fill={rgb,255:red,227;green,243;blue,153}] (5.500,-2.160) rectangle (7.400,-3.040);
\node[text=black,align=center,font=\scriptsize\sffamily,inner sep=0pt] at (6.450,-2.600)
  {198/220\\90\%};
\fill[fill={rgb,255:red,57;green,167;blue,88}] (7.400,-2.160) rectangle (9.300,-3.040);
\node[text=white,align=center,font=\scriptsize\sffamily,inner sep=0pt] at (8.350,-2.600)
  {150/150\\100\%};
\fill[fill={rgb,255:red,122;green,198;blue,101}] (9.300,-2.160) rectangle (11.200,-3.040);
\node[text=black,align=center,font=\scriptsize\sffamily,inner sep=0pt] at (10.250,-2.600)
  {155/160\\97\%};
\fill[fill={rgb,255:red,227;green,243;blue,153}] (11.200,-2.160) rectangle (13.100,-3.040);
\node[text=black,align=center,font=\scriptsize\sffamily,inner sep=0pt] at (12.150,-2.600)
  {171/190\\90\%};
\fill[fill={rgb,255:red,227;green,243;blue,153}] (13.100,-2.160) rectangle (15.000,-3.040);
\node[text=black,align=center,font=\scriptsize\sffamily,inner sep=0pt] at (14.050,-2.600)
  {135/150\\90\%};
\node[anchor=west,font=\small\sffamily,inner sep=3pt] at (15.120,-2.600)
  {$\!1004$};
\draw[blue!60!black,line width=1.5pt] (7.400,-2.160) rectangle (9.300,-3.040);
\draw[blue!60!black,line width=1.5pt] (13.100,-2.160) rectangle (15.000,-3.040);
\fill[black!5] (0,-3.040) rectangle (3.600,-3.920);
\node[anchor=west,font=\small\sffamily,inner sep=2pt] at (0.10,-3.480) {MediTab};
\fill[fill={rgb,255:red,57;green,167;blue,88}] (3.600,-3.040) rectangle (5.500,-3.920);
\node[text=white,align=center,font=\scriptsize\sffamily,inner sep=0pt] at (4.550,-3.480)
  {200/200\\100\%};
\fill[fill={rgb,255:red,57;green,167;blue,88}] (5.500,-3.040) rectangle (7.400,-3.920);
\node[text=white,align=center,font=\scriptsize\sffamily,inner sep=0pt] at (6.450,-3.480)
  {220/220\\100\%};
\fill[fill={rgb,255:red,57;green,167;blue,88}] (7.400,-3.040) rectangle (9.300,-3.920);
\node[text=white,align=center,font=\scriptsize\sffamily,inner sep=0pt] at (8.350,-3.480)
  {150/150\\100\%};
\fill[fill={rgb,255:red,237;green,95;blue,60}] (9.300,-3.040) rectangle (11.200,-3.920);
\node[text=white,align=center,font=\scriptsize\sffamily,inner sep=0pt] at (10.250,-3.480)
  {122/160\\76\%};
\fill[fill={rgb,255:red,227;green,243;blue,153}] (11.200,-3.040) rectangle (13.100,-3.920);
\node[text=black,align=center,font=\scriptsize\sffamily,inner sep=0pt] at (12.150,-3.480)
  {171/190\\90\%};
\fill[fill={rgb,255:red,227;green,243;blue,153}] (13.100,-3.040) rectangle (15.000,-3.920);
\node[text=black,align=center,font=\scriptsize\sffamily,inner sep=0pt] at (14.050,-3.480)
  {135/150\\90\%};
\node[anchor=west,font=\small\sffamily,inner sep=3pt] at (15.120,-3.480)
  {$\!998$};
\draw[blue!60!black,line width=1.5pt] (3.600,-3.040) rectangle (5.500,-3.920);
\draw[blue!60!black,line width=1.5pt] (5.500,-3.040) rectangle (7.400,-3.920);
\draw[blue!60!black,line width=1.5pt] (7.400,-3.040) rectangle (9.300,-3.920);
\draw[blue!60!black,line width=1.5pt] (13.100,-3.040) rectangle (15.000,-3.920);
\fill[black!5] (0,-3.920) rectangle (3.600,-4.800);
\node[anchor=west,font=\small\sffamily,inner sep=2pt] at (0.10,-4.360) {TAPTAP};
\fill[fill={rgb,255:red,57;green,167;blue,88}] (3.600,-3.920) rectangle (5.500,-4.800);
\node[text=white,align=center,font=\scriptsize\sffamily,inner sep=0pt] at (4.550,-4.360)
  {200/200\\100\%};
\fill[fill={rgb,255:red,57;green,167;blue,88}] (5.500,-3.920) rectangle (7.400,-4.800);
\node[text=white,align=center,font=\scriptsize\sffamily,inner sep=0pt] at (6.450,-4.360)
  {220/220\\100\%};
\fill[fill={rgb,255:red,57;green,167;blue,88}] (7.400,-3.920) rectangle (9.300,-4.800);
\node[text=white,align=center,font=\scriptsize\sffamily,inner sep=0pt] at (8.350,-4.360)
  {150/150\\100\%};
\fill[fill={rgb,255:red,208;green,41;blue,39}] (9.300,-3.920) rectangle (11.200,-4.800);
\node[text=white,align=center,font=\scriptsize\sffamily,inner sep=0pt] at (10.250,-4.360)
  {117/160\\73\%};
\fill[fill={rgb,255:red,227;green,243;blue,153}] (11.200,-3.920) rectangle (13.100,-4.800);
\node[text=black,align=center,font=\scriptsize\sffamily,inner sep=0pt] at (12.150,-4.360)
  {171/190\\90\%};
\fill[fill={rgb,255:red,227;green,243;blue,153}] (13.100,-3.920) rectangle (15.000,-4.800);
\node[text=black,align=center,font=\scriptsize\sffamily,inner sep=0pt] at (14.050,-4.360)
  {135/150\\90\%};
\node[anchor=west,font=\small\sffamily,inner sep=3pt] at (15.120,-4.360)
  {$\!993$};
\draw[blue!60!black,line width=1.5pt] (3.600,-3.920) rectangle (5.500,-4.800);
\draw[blue!60!black,line width=1.5pt] (5.500,-3.920) rectangle (7.400,-4.800);
\draw[blue!60!black,line width=1.5pt] (7.400,-3.920) rectangle (9.300,-4.800);
\draw[blue!60!black,line width=1.5pt] (13.100,-3.920) rectangle (15.000,-4.800);
\fill[black!5] (0,-4.800) rectangle (3.600,-5.680);
\node[anchor=west,font=\small\sffamily,inner sep=2pt] at (0.10,-5.240) {TabDPT};
\fill[fill={rgb,255:red,157;green,213;blue,105}] (3.600,-4.800) rectangle (5.500,-5.680);
\node[text=black,align=center,font=\scriptsize\sffamily,inner sep=0pt] at (4.550,-5.240)
  {190/200\\95\%};
\fill[fill={rgb,255:red,227;green,243;blue,153}] (5.500,-4.800) rectangle (7.400,-5.680);
\node[text=black,align=center,font=\scriptsize\sffamily,inner sep=0pt] at (6.450,-5.240)
  {198/220\\90\%};
\fill[fill={rgb,255:red,183;green,224;blue,117}] (7.400,-4.800) rectangle (9.300,-5.680);
\node[text=black,align=center,font=\scriptsize\sffamily,inner sep=0pt] at (8.350,-5.240)
  {140/150\\93\%};
\fill[fill={rgb,255:red,122;green,198;blue,101}] (9.300,-4.800) rectangle (11.200,-5.680);
\node[text=black,align=center,font=\scriptsize\sffamily,inner sep=0pt] at (10.250,-5.240)
  {155/160\\97\%};
\fill[fill={rgb,255:red,227;green,243;blue,153}] (11.200,-4.800) rectangle (13.100,-5.680);
\node[text=black,align=center,font=\scriptsize\sffamily,inner sep=0pt] at (12.150,-5.240)
  {171/190\\90\%};
\fill[fill={rgb,255:red,227;green,243;blue,153}] (13.100,-4.800) rectangle (15.000,-5.680);
\node[text=black,align=center,font=\scriptsize\sffamily,inner sep=0pt] at (14.050,-5.240)
  {135/150\\90\%};
\node[anchor=west,font=\small\sffamily,inner sep=3pt] at (15.120,-5.240)
  {$\!989$};
\draw[blue!60!black,line width=1.5pt] (13.100,-4.800) rectangle (15.000,-5.680);
\fill[black!5] (0,-5.680) rectangle (3.600,-6.560);
\node[anchor=west,font=\small\sffamily,inner sep=2pt] at (0.10,-6.120) {XTab};
\fill[fill={rgb,255:red,57;green,167;blue,88}] (3.600,-5.680) rectangle (5.500,-6.560);
\node[text=white,align=center,font=\scriptsize\sffamily,inner sep=0pt] at (4.550,-6.120)
  {200/200\\100\%};
\fill[fill={rgb,255:red,227;green,243;blue,153}] (5.500,-5.680) rectangle (7.400,-6.560);
\node[text=black,align=center,font=\scriptsize\sffamily,inner sep=0pt] at (6.450,-6.120)
  {198/220\\90\%};
\fill[fill={rgb,255:red,227;green,243;blue,153}] (7.400,-5.680) rectangle (9.300,-6.560);
\node[text=black,align=center,font=\scriptsize\sffamily,inner sep=0pt] at (8.350,-6.120)
  {135/150\\90\%};
\fill[fill={rgb,255:red,227;green,243;blue,153}] (9.300,-5.680) rectangle (11.200,-6.560);
\node[text=black,align=center,font=\scriptsize\sffamily,inner sep=0pt] at (10.250,-6.120)
  {144/160\\90\%};
\fill[fill={rgb,255:red,227;green,243;blue,153}] (11.200,-5.680) rectangle (13.100,-6.560);
\node[text=black,align=center,font=\scriptsize\sffamily,inner sep=0pt] at (12.150,-6.120)
  {171/190\\90\%};
\fill[fill={rgb,255:red,227;green,243;blue,153}] (13.100,-5.680) rectangle (15.000,-6.560);
\node[text=black,align=center,font=\scriptsize\sffamily,inner sep=0pt] at (14.050,-6.120)
  {135/150\\90\%};
\node[anchor=west,font=\small\sffamily,inner sep=3pt] at (15.120,-6.120)
  {$\!983$};
\draw[blue!60!black,line width=1.5pt] (3.600,-5.680) rectangle (5.500,-6.560);
\draw[blue!60!black,line width=1.5pt] (13.100,-5.680) rectangle (15.000,-6.560);
\fill[black!5] (0,-6.560) rectangle (3.600,-7.440);
\node[anchor=west,font=\small\sffamily,inner sep=2pt] at (0.10,-7.000) {SAINT};
\fill[fill={rgb,255:red,110;green,192;blue,100}] (3.600,-6.560) rectangle (5.500,-7.440);
\node[text=black,align=center,font=\scriptsize\sffamily,inner sep=0pt] at (4.550,-7.000)
  {195/200\\98\%};
\fill[fill={rgb,255:red,227;green,243;blue,153}] (5.500,-6.560) rectangle (7.400,-7.440);
\node[text=black,align=center,font=\scriptsize\sffamily,inner sep=0pt] at (6.450,-7.000)
  {198/220\\90\%};
\fill[fill={rgb,255:red,125;green,199;blue,101}] (7.400,-6.560) rectangle (9.300,-7.440);
\node[text=black,align=center,font=\scriptsize\sffamily,inner sep=0pt] at (8.350,-7.000)
  {145/150\\97\%};
\fill[fill={rgb,255:red,208;green,41;blue,39}] (9.300,-6.560) rectangle (11.200,-7.440);
\node[text=white,align=center,font=\scriptsize\sffamily,inner sep=0pt] at (10.250,-7.000)
  {117/160\\73\%};
\fill[fill={rgb,255:red,57;green,167;blue,88}] (11.200,-6.560) rectangle (13.100,-7.440);
\node[text=white,align=center,font=\scriptsize\sffamily,inner sep=0pt] at (12.150,-7.000)
  {190/190\\100\%};
\fill[fill={rgb,255:red,227;green,243;blue,153}] (13.100,-6.560) rectangle (15.000,-7.440);
\node[text=black,align=center,font=\scriptsize\sffamily,inner sep=0pt] at (14.050,-7.000)
  {135/150\\90\%};
\node[anchor=west,font=\small\sffamily,inner sep=3pt] at (15.120,-7.000)
  {$\!980$};
\draw[blue!60!black,line width=1.5pt] (11.200,-6.560) rectangle (13.100,-7.440);
\draw[blue!60!black,line width=1.5pt] (13.100,-6.560) rectangle (15.000,-7.440);

\fill[fill={rgb,255:red,171;green,6;blue,38}] (16.600,-7.440) rectangle (16.980,-7.254);
\fill[fill={rgb,255:red,183;green,17;blue,38}] (16.600,-7.254) rectangle (16.980,-7.068);
\fill[fill={rgb,255:red,196;green,30;blue,39}] (16.600,-7.068) rectangle (16.980,-6.882);
\fill[fill={rgb,255:red,208;green,41;blue,39}] (16.600,-6.882) rectangle (16.980,-6.696);
\fill[fill={rgb,255:red,218;green,54;blue,42}] (16.600,-6.696) rectangle (16.980,-6.510);
\fill[fill={rgb,255:red,226;green,71;blue,49}] (16.600,-6.510) rectangle (16.980,-6.324);
\fill[fill={rgb,255:red,233;green,85;blue,56}] (16.600,-6.324) rectangle (16.980,-6.138);
\fill[fill={rgb,255:red,241;green,102;blue,64}] (16.600,-6.138) rectangle (16.980,-5.952);
\fill[fill={rgb,255:red,245;green,117;blue,71}] (16.600,-5.952) rectangle (16.980,-5.766);
\fill[fill={rgb,255:red,247;green,132;blue,78}] (16.600,-5.766) rectangle (16.980,-5.580);
\fill[fill={rgb,255:red,250;green,150;blue,86}] (16.600,-5.580) rectangle (16.980,-5.394);
\fill[fill={rgb,255:red,252;green,165;blue,93}] (16.600,-5.394) rectangle (16.980,-5.208);
\fill[fill={rgb,255:red,253;green,181;blue,103}] (16.600,-5.208) rectangle (16.980,-5.022);
\fill[fill={rgb,255:red,253;green,193;blue,113}] (16.600,-5.022) rectangle (16.980,-4.836);
\fill[fill={rgb,255:red,254;green,204;blue,123}] (16.600,-4.836) rectangle (16.980,-4.650);
\fill[fill={rgb,255:red,254;green,218;blue,134}] (16.600,-4.650) rectangle (16.980,-4.464);
\fill[fill={rgb,255:red,254;green,228;blue,145}] (16.600,-4.464) rectangle (16.980,-4.278);
\fill[fill={rgb,255:red,254;green,236;blue,159}] (16.600,-4.278) rectangle (16.980,-4.092);
\fill[fill={rgb,255:red,255;green,243;blue,172}] (16.600,-4.092) rectangle (16.980,-3.906);
\fill[fill={rgb,255:red,255;green,251;blue,184}] (16.600,-3.906) rectangle (16.980,-3.720);
\fill[fill={rgb,255:red,250;green,253;blue,184}] (16.600,-3.720) rectangle (16.980,-3.534);
\fill[fill={rgb,255:red,241;green,249;blue,172}] (16.600,-3.534) rectangle (16.980,-3.348);
\fill[fill={rgb,255:red,230;green,245;blue,157}] (16.600,-3.348) rectangle (16.980,-3.162);
\fill[fill={rgb,255:red,221;green,241;blue,145}] (16.600,-3.162) rectangle (16.980,-2.976);
\fill[fill={rgb,255:red,211;green,236;blue,135}] (16.600,-2.976) rectangle (16.980,-2.790);
\fill[fill={rgb,255:red,197;green,230;blue,126}] (16.600,-2.790) rectangle (16.980,-2.604);
\fill[fill={rgb,255:red,185;green,225;blue,118}] (16.600,-2.604) rectangle (16.980,-2.418);
\fill[fill={rgb,255:red,171;green,219;blue,109}] (16.600,-2.418) rectangle (16.980,-2.232);
\fill[fill={rgb,255:red,157;green,213;blue,105}] (16.600,-2.232) rectangle (16.980,-2.046);
\fill[fill={rgb,255:red,142;green,207;blue,103}] (16.600,-2.046) rectangle (16.980,-1.860);
\fill[fill={rgb,255:red,125;green,199;blue,101}] (16.600,-1.860) rectangle (16.980,-1.674);
\fill[fill={rgb,255:red,110;green,192;blue,100}] (16.600,-1.674) rectangle (16.980,-1.488);
\fill[fill={rgb,255:red,90;green,183;blue,96}] (16.600,-1.488) rectangle (16.980,-1.302);
\fill[fill={rgb,255:red,72;green,174;blue,92}] (16.600,-1.302) rectangle (16.980,-1.116);
\fill[fill={rgb,255:red,54;green,166;blue,87}] (16.600,-1.116) rectangle (16.980,-0.930);
\fill[fill={rgb,255:red,33;green,156;blue,82}] (16.600,-0.930) rectangle (16.980,-0.744);
\fill[fill={rgb,255:red,22;green,145;blue,77}] (16.600,-0.744) rectangle (16.980,-0.558);
\fill[fill={rgb,255:red,15;green,132;blue,70}] (16.600,-0.558) rectangle (16.980,-0.372);
\fill[fill={rgb,255:red,9;green,121;blue,64}] (16.600,-0.372) rectangle (16.980,-0.186);
\fill[fill={rgb,255:red,3;green,110;blue,58}] (16.600,-0.186) rectangle (16.980,-0.000);
\draw[black,line width=0.5pt] (16.600,-7.440) rectangle (16.980,0.000);
\draw[black,line width=0.5pt] (16.980,-7.440)--(17.130,-7.440);
\node[anchor=west,font=\tiny\sffamily] at (17.160,-7.440) {70\%};
\draw[black,line width=0.5pt] (16.980,-6.377)--(17.130,-6.377);
\node[anchor=west,font=\tiny\sffamily] at (17.160,-6.377) {75\%};
\draw[black,line width=0.5pt] (16.980,-5.314)--(17.130,-5.314);
\node[anchor=west,font=\tiny\sffamily] at (17.160,-5.314) {80\%};
\draw[black,line width=0.5pt] (16.980,-4.251)--(17.130,-4.251);
\node[anchor=west,font=\tiny\sffamily] at (17.160,-4.251) {85\%};
\draw[black,line width=0.5pt] (16.980,-3.189)--(17.130,-3.189);
\node[anchor=west,font=\tiny\sffamily] at (17.160,-3.189) {90\%};
\draw[black,line width=0.5pt] (16.980,-2.126)--(17.130,-2.126);
\node[anchor=west,font=\tiny\sffamily] at (17.160,-2.126) {95\%};
\draw[black,line width=0.5pt] (16.980,-1.063)--(17.130,-1.063);
\node[anchor=west,font=\tiny\sffamily] at (17.160,-1.063) {100\%};
\node[rotate=90,anchor=south,font=\scriptsize\sffamily] at (17.880,-3.720) {Score / max possible};

\end{tikzpicture}
\caption{Metric-layer performance profile under Heart Failure use-case weights, computed from the same independent-literature-based property audit (Section~\ref{sec:audit-note-v2}). Each cell shows the raw metric score relative to the use-case-adjusted maximum and the corresponding percentage. Blue borders mark the highest-scoring model in each metric column. The same eight models appear as in Figure~\ref{fig:id_heatmap}; comparing the two figures shows the ranking reorders within the group rather than a fixed top group holding constant. $\mathcal{R}$ is the overall weighted ranking score (max\,=\,1070).}
\label{fig:hf_heatmap}
\end{figure}
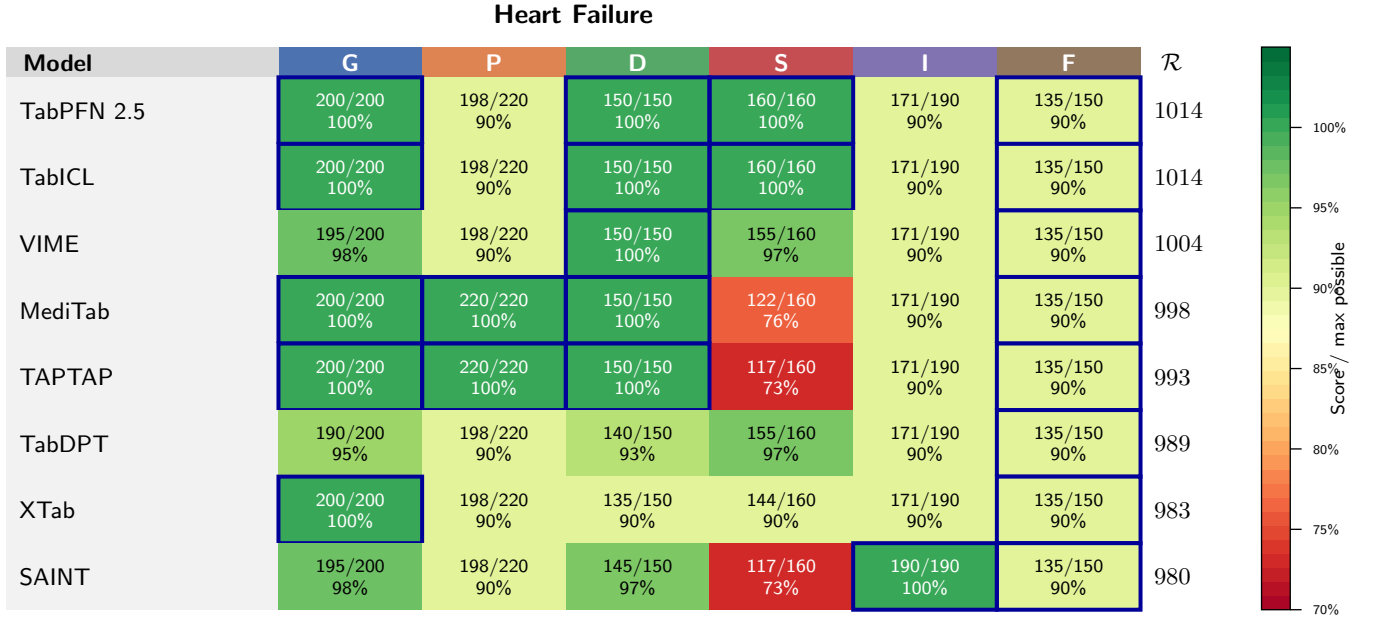

Applying these sub-metric weights to the scores in Figure~\ref{fig:tfm_scores} produces the metric-layer profile and the overall ranking together. Figure~\ref{fig:hf_heatmap} shows the metric-layer profile visually: compared to Figure~\ref{fig:id_heatmap}, the shift in clinical priorities produces a markedly different pattern of strengths and shortfalls, with scalability now carrying more weight and data-efficient learning carrying less. The six columns $\mathcal{G}$, $\mathcal{P}$, $\mathcal{D}$, $\mathcal{S}$, $\mathcal{I}$, and $\mathcal{F}$ in Table~\ref{tab:hf_ranking} give the same scores numerically. The final column $\mathcal{R}(\tau_i)$ is the integrated overall score. Reading the table column by column before reading the final ranking is often more informative, because a model that is uniformly strong across six dimensions is a fundamentally different candidate from one that maximizes the total by excelling on a few dimensions while being weak on others.

\begin{table}[t]
\centering
\small
\caption{Model ranking for the heart failure use case, computed from the audited property assignments (Section~\ref{sec:audit-note}). Scores represent the weighted sum $\mathcal{R}(\tau_i)$ across all sub-metrics under the heart failure sub-metric weights. The maximum possible score is 1070. The same eight models appear as in Table~\ref{tab:id_ranking}; comparing the two tables shows the ranking reorders within the group -- most notably, SAINT falls from sixth in the iron deficiency table to last here, as the heavier heart failure scalability weighting exposes its independently-documented resource-utilization weakness, while TabDPT and XTab, whose scalability scores were not revised by this audit, close the gap and pass it -- rather than a fixed top group holding constant regardless of clinical priority.}
\label{tab:hf_ranking}
\begin{tabular}{|l|c|c|c|c|c|c|c|}
\hline
\textbf{Model} & $\mathcal{G}$ & $\mathcal{P}$ & $\mathcal{D}$ & $\mathcal{S}$ & $\mathcal{I}$ & $\mathcal{F}$ & $\mathcal{R}(\tau_i)$ \\
\hline
Max possible    & 200 & 220 & 150 & 160 & 190 & 150 & 1070 \\ \hline
TabPFN~2.5      & 200 & 198 & 150 & 160 & 171 & 135 & \textbf{1014} \\
TabICL          & 200 & 198 & 150 & 160 & 171 & 135 & \textbf{1014} \\
VIME            & 195 & 198 & 150 & 155 & 171 & 135 & 1004 \\
MediTab         & 200 & 220 & 150 & 122 & 171 & 135 &  998 \\
TAPTAP          & 200 & 220 & 150 & 117 & 171 & 135 &  993 \\
TabDPT          & 190 & 198 & 140 & 155 & 171 & 135 &  989 \\
XTab            & 200 & 198 & 135 & 144 & 171 & 135 &  983 \\
SAINT           & 195 & 198 & 145 & 117 & 190 & 135 &  980 \\
\hline
\end{tabular}
\end{table}

TabPFN~2.5 and TabICL again tie for first, now at 1014/1070, sharing the same identical sub-metric profile discussed in Section~\ref{sec:id-usecase}. VIME rises to third at 1004: heart failure's heavier scalability weighting (160/1070, versus 110/1020 in iron deficiency) rewards VIME's high, unrevised scalability score (155/160) relative to models whose scalability sub-metrics this audit revised downward. MediTab (998) and TAPTAP (993) again reach the privacy ceiling (220/220), but their scalability scores (122/160 and 117/160 respectively) are a larger absolute penalty here than in the iron deficiency case, since heart failure weights scalability more heavily; TAPTAP's place in this table again carries the same contested-privacy caveat discussed in Section~\ref{sec:id-usecase}.

TabDPT (989) and XTab (983) both move ahead of SAINT (980, last in this table) specifically because SAINT's resource-utilization and embedding-generation properties were independently found to be markedly weaker than its own paper implies -- the TabZilla finding that SAINT's training time is the slowest of 19 tested algorithms (Section~\ref{sec:id-usecase}) -- while TabDPT's and XTab's scalability profiles were not challenged by any independent source found during this audit and so remain at their prior, primary-source-only values (155/160 and 144/160 respectively). This is a purely relative reordering for TabDPT and XTab: neither model's own scores changed under this audit, and their rise past SAINT reflects SAINT's documented weakness becoming more costly under heavier scalability weighting, not new evidence in either model's favor.

\begin{table}[t]
\centering
\small
\caption{Super-metric scores for the heart failure use case. TLC = Transfer Learning Capacity ($\mathcal{TLC}$, max 370), MR = Model Resilience ($\mathcal{MR}$, max 190), EPT = Efficiency-Privacy Trade-off ($\mathcal{EPT}$, max 530), CT = Clinical Trustworthiness ($\mathcal{CT}$, max 340). Scores are weighted sums using the heart failure sub-metric weights.}
\label{tab:hf_supermetrics}
\begin{tabular}{|l|c|c|c|c|}
\hline
\textbf{Model} & \textbf{TLC} (370) & \textbf{MR} (190) & \textbf{EPT} (530) & \textbf{CT} (340) \\
\hline
TabPFN~2.5      & 361 & 185 & 508 & 306 \\
TabICL          & 361 & 185 & 508 & 306 \\
VIME            & 356 & 180 & 503 & 306 \\
MediTab         & 370 & 190 & 492 & 306 \\
TAPTAP          & 370 & 190 & 487 & 306 \\
TabDPT          & 341 & 165 & 493 & 306 \\
XTab            & 353 & 180 & 477 & 306 \\
SAINT           & 351 & 175 & 460 & 325 \\
\hline
\end{tabular}
\end{table}

On Transfer Learning Capacity, MediTab and TAPTAP lead at the ceiling (370/370), combining maximal generalizability with a perfect privacy profile. TabPFN~2.5 and TabICL follow at 361, one point per privacy sub-metric short of the ceiling. VIME reaches 356, XTab 353, and SAINT 351; TabDPT trails at 341, the same schema-robustness and data-noise-robustness weakness noted in Section~\ref{sec:id-usecase}.

Model Resilience: MediTab and TAPTAP lead at the ceiling (190/190), TabPFN~2.5 and TabICL follow at 185, VIME and XTab tie at 180, SAINT trails at 175, and TabDPT is lowest at 165.

The Efficiency-Privacy Trade-off reaches a higher maximum in the heart failure case (530, versus 500 in iron deficiency) because the higher scalability weights increase the total contribution of the S sub-metrics included in EPT. TabPFN~2.5 and TabICL lead at 508. VIME follows at 503, TabDPT at 493, MediTab at 492, TAPTAP at 487, and XTab at 477; SAINT trails at 460, the lowest in the table, penalized by the same independently-documented resource-utilization weakness discussed above.

Clinical Trustworthiness again splits the table along a single line: SAINT alone reaches 325/340 through its documented interpretability mechanism, while the remaining seven models score 306/340.

For a heart failure prediction program, \textbf{TabPFN~2.5 and TabICL are again the joint primary recommendation}, tied for the overall ranking at 1014/1070 with the same identical sub-metric profile discussed in Section~\ref{sec:id-usecase}. VIME is the strongest alternative here specifically because its scalability profile was not revised downward by this audit and holds up well under heart failure's heavier scalability weighting; MediTab and TAPTAP remain the strongest privacy-ceiling alternatives, and SAINT, despite reaching the interpretability ceiling that the co-leaders do not, falls to last place in this table as the heavier scalability weighting exposes its independently-documented resource-utilization weakness.

Comparing the two use cases makes the framework's adaptive value concrete. TabPFN~2.5 and TabICL lead both rankings as an exact tie, but the models immediately behind them reorder: SAINT falls from sixth in the iron deficiency ranking to last among the top eight in the heart failure ranking, specifically because heart failure's heavier scalability weighting turns an independently-documented resource-utilization weakness (Section~\ref{sec:id-usecase}) into a larger absolute penalty, while TabDPT and XTab, whose scalability profiles were not challenged by any independent source found during this audit, close the gap and pass it. This is the kind of mechanism the framework's methodology (Section~\ref{sec:architecture}) was designed to surface, and Section~\ref{sec:audit-note} discusses how this second, independent-literature-based audit pass changed which models demonstrate it.

\subsection{Note on the Property Audit and Its Effect on the Ranking}
\label{sec:audit-note}

The sub-metric scores used throughout this section differ from those in an earlier draft of this work, in which four models (FT-Transformer, MediTab, TabNet, and a duplicated, over-credited row for TabTransformer) ranked first through fourth under both the iron deficiency and heart failure weightings, and every other surveyed model, including the TabPFN family, was mathematically incapable of reaching the top of the ranking under any combination of sub-metric weights. A full re-derivation of the property assignments in Table~\ref{tab:drawback_properties} against each model's original publication, rather than the mixed literature-and-holistic-judgment process used previously, found that this was a property-assignment artifact rather than a genuine finding about model quality: FT-Transformer's original paper explicitly states it uses no pretraining, and several of the four originally-leading models had been credited with properties (a pretraining phase, missing-data handling, differential-privacy compliance) that their own publications do not claim, while the TabPFN family and several other transfer/pretraining models had been under-credited on properties central to their own documented contributions, most notably data-efficient and zero-shot learning for TabPFN and its successors.

Under the corrected assignments, the resulting scores are close to the maximum for a wider set of models, all four categories of the model taxonomy (prediction-focused, transfer/pretraining, generative, and LLM- or NLP-oriented) contain at least one model within reach of the top eight, and the two case studies produce different orderings among the top group rather than the same four models regardless of priority. Of the 45 models surveyed, 41 remain unable to win outright under any sub-metric weighting because they are pointwise dominated by another single model on all 25 sub-metrics -- a level of differentiation among the small number of undominated models (TabPFN~v2, TabICL, and MediTab, under this first re-derivation's scores) that is expected given how many models in the survey are architecturally similar variants of a small number of design paradigms, rather than a sign that the underlying instrument still lacks resolution. This dominance analysis was not repeated on the second, independent-literature-based re-derivation described below, since several of the models it names (TabPFN~v2 in particular) no longer occupy the position the analysis describes; it is reported here as a property of the first re-derivation's output rather than a claim that still holds today. Four entries in the extended model list (P-Transformer, DTT, GPT4Table, and the SPROUT/UniTTab pair) could not be matched to a verifiable primary source under their listed identity during the re-derivation; their scores are placeholders pending correction or removal, and no case-study conclusion in this section depends on any of the four. We report this correction in the interest of the same transparency and auditability the framework is designed to provide: a scoring instrument whose output survives a re-derivation from primary sources is a stronger basis for clinical decision support than one that does not, and the exercise of re-deriving it here is itself evidence for the property-based scoring approach (Section~\ref{sec:architecture}) over an unaudited alternative.

\subsubsection{A Second Audit Pass: Independent Literature in Place of Primary Sources}
\label{sec:audit-note-v2}

The property assignments described above were re-derived a second time after this paper's author identified a structural limitation in the first re-derivation: scoring each model's properties against its own publication alone systematically rewards models whose papers compare against a narrow or dated set of baselines, and penalizes models whose papers are comparatively candid about their own limitations. FT-Transformer's paper is the clearest example: it compares itself only to ResNet, MLP, and XGBoost, so its resource-utilization properties were held to an ambiguous, 2-of-3-absent rating that made it look inefficient relative to its actual peer set of tabular deep-learning architectures, not because independent evidence supported that rating but because no independent evidence was consulted at all. This second pass re-derives every property assignment against independent comparison and survey literature -- third-party benchmarks, replication studies, and surveys that evaluate each model against its true competitors -- rather than against the model's own publication, and keeps the first pass's value wherever no independent evidence addressing a given model and property was found. Sixteen of the 45 surveyed models have at least one sub-metric revised under this process; the remaining 28, including four of the eight models in the first pass's top eight (TabICL, SAINT, VIME, MediTab), are unchanged. Figure~\ref{fig:tfm_scores}, and Tables~\ref{tab:id_ranking}--\ref{tab:hf_supermetrics} all reflect the revised scores throughout this section.

The clearest confirmation of the motivating premise is FT-Transformer itself: benchmarked against its true deep-learning peer set (McElfresh et al., ``TabZilla,'' NeurIPS 2023 D\&B) rather than ResNet, MLP, and XGBoost alone, its training time is faster than SAINT, NODE, DANet, and TabNet, and its resource-utilization sub-metrics rise from 8 to the ceiling of 10. A follow-up check, run specifically for FT-Transformer at the paper author's request rather than as part of the systematic pass across all 45 models, found the same pattern on a second property: FT-Transformer's own paper states plainly that it uses no pretraining, but two independent sources -- XTab (Zhu et al., ICML 2023), which plugs FT-Transformer in as a backbone and reports that XTab-pretrained FT-Transformer beats other state-of-the-art tabular deep-learning models, and a production case study at Booking.com (arXiv:2405.13692) reporting an eight-percent relative gain from self-supervised pretraining plus fine-tuning over training FT-Transformer from scratch -- show that its specific architecture achieves a real transfer benefit under external pretraining frameworks built by other researchers. Its Built-in Pre-Training property is accordingly flipped to present (IDT, CDT, FAC, and ZIA rise from 9 to 10; PRD and FLC rise from 8 to 9), raising FT-Transformer's combined rank from roughly thirty-first to fifteenth of 45, still short of the new top-eight threshold. Because this specific check was run only for FT-Transformer and not for the other roughly fifteen to twenty models in the survey currently marked Built-in Pre-Training absent, some of which may have an equally applicable independent pretraining-framework result that simply was not searched for, this is reported as a known, deliberate gap in this audit's consistency rather than a claim that FT-Transformer was singled out on the merits alone. A second follow-up check, again run specifically for FT-Transformer, revisited its Data Noise Robustness sub-metric against the same TabZilla-adjacent literature already consulted for SAINT, TabNet, and NODE below: Grinsztajn, Oyallon, and Varoquaux (NeurIPS 2022 D\&B) find that FT-Transformer, unlike MLP-style networks, is not badly hurt by the uninformative or noisy features common in tabular data, because its Feature Tokenizer, like a tree-based split, is a pointwise rather than rotation-invariant operation -- the same structural property the paper identifies as separating tree-based models from noise-vulnerable neural architectures. Its Handle Noisy Inputs property is accordingly flipped to present, raising Data Noise Robustness from 8 to 9 and its combined rank to eleventh of 45, still short of the top-eight threshold; a parallel check of its Data Augmentation Capability found no comparably specific independent evidence and left that property unchanged. The same benchmark runs in the opposite direction for three prediction-focused models previously credited with efficient resource utilization on an ambiguous or unchallenged reading of their own papers: SAINT, TabNet, and NODE are directly benchmarked in TabZilla and found among the slowest of the 19 algorithms tested (SAINT slowest overall, NODE second-slowest), and their resource-utilization sub-metrics fall accordingly.

The headline change, however, runs in the opposite direction from the audit's founding motivation: it is TabPFN~v2's own paper, not a paper describing a weaker or less-benchmarked model, that turns out to have made the model look better than the independent literature now supports. Three independent findings converge specifically on TabPFN~v2. The TabICL paper itself \cite{qu2025tabicl}, an independent benchmark rather than TabPFN~v2's own report, found that TabPFN~v2 requires subsampling to 30,000 rows to avoid out-of-memory errors and is roughly ten times slower than TabICL on large, many-column datasets, dropping its resource-utilization, computational-scalability, and training-time-efficiency sub-metrics from 10 to 7 and its feature-scalability sub-metric from 9 to 8. Four independent interpretability-probing papers published in 2026 (arXiv:2601.08181, arXiv:2601.23068, arXiv:2602.02162, arXiv:2603.29946) found that TabPFN~v2 has no native interpretability mechanism and depends entirely on post-hoc tooling, contradicting the SHAP/LOCO framing of its own Nature paper and dropping its four interpretability sub-metrics from 10 to 9. TabFSBench (Wu et al., ICML 2025), the first benchmark purpose-built for feature-shift testing, found TabPFN~v2 among the models with limited applicability under feature shift, dropping its schema-robustness sub-metric from 10 to 9. No single one of these findings would have been enough to unseat TabPFN~v2 on its own, but together they cost it 48 points under the iron deficiency weights (981 to 933) and 62 points under the heart failure weights (1028 to 966), moving it from an outright first-place finish in both case studies to twelfth of 45 (iron deficiency) and seventeenth of 45 (heart failure). TabPFN (v1) and TabNet fall out of the top eight for closely related reasons: TabPFN is directly implicated by the same TabZilla resource-utilization finding and by an analogous TabFSBench schema-robustness finding that both apply to the TabPFN lineage, and TabNet's own TabZilla training-time figure -- the median of the 19 algorithms tested -- is no longer reconcilable with a perfect resource-utilization score.

TabPFN~2.5 and TabICL, by contrast, pick up no independent penalty of this kind and are now tied for first place in both case studies with byte-identical 25-sub-metric profiles. This is not entirely a sign that the two models are without weaknesses: no independent literature was found, as of this audit, probing either model's scalability or interpretability claims the way the TabICL paper and the four interpretability papers probed TabPFN~v2, which is at least partly a consequence of both models being new enough (2025) that comprehensive independent scrutiny has not yet accumulated. The one point of difference the two models previously had, TabPFN~2.5's Data Noise Robustness (9 versus TabICL's 10), closed after TabArena (Erickson et al., NeurIPS 2025 D\&B) reported that TabPFN-2.5 substantially outperforms tuned tree-based models on a broad, realistic benchmark suite, consistent with the robust-scaling and soft-clipping mechanisms its own technical report describes; this specific flip is reported at moderate rather than high confidence, since TabArena measures aggregate rank rather than an isolated noise-injection ablation.

TAPTAP, TabDPT, and XTab enter the top eight in place of TabPFN~v2, TabPFN, and TabNet. TabDPT's and XTab's own scores are entirely unchanged by this audit -- no independent literature addressing either model was found in any of the six domain files reviewed -- so their rise is a purely relative consequence of stronger peers falling, not new positive evidence about either model; this is flagged here as a genuine limitation of the current evidence base and not treated as equivalent to TabPFN~2.5's more thoroughly cross-checked standing. TAPTAP's rise rests partly on a tentative, low-confidence Handle Missing Data flip (shared with MediTab, and separately confirmed not to be selection-determining on its own) and partly, more materially, on a genuinely unresolved question: an independent study of memorization in GPT-2-derived tabular generators found measurable numeric-string leakage in the plain GPT-2 backbone that TAPTAP continues pretraining from, an indirect challenge to the ``present'' Data Anonymization Techniques rating that TAPTAP's own paper claims for itself. Because this specific challenge is indirect -- TAPTAP itself was not directly attacked in that study -- and because a separate, more direct contradiction exists elsewhere in the privacy literature for a different model (REaLTabFormer, where two independent attack studies using different methodologies reach opposite conclusions about its privacy risk, and which this audit likewise declines to resolve), this audit's conflict-resolution rule retains TAPTAP's prior ``present'' value rather than resolving the question either way. Readers should treat TAPTAP's place in the top eight as conditional on this open question: resolving it toward ``absent,'' as the underlying evidence would argue, drops TAPTAP's combined ranking below XTab's and out of the top eight, with TabPFN~3 taking its place. This caveat is reported for transparency, consistent with the same auditability principle that motivated re-deriving the scores in the first place, and is not something this pass resolves unilaterally.

The first pass's broader conclusion -- that the corrected scores spread near-maximal performance across a wider set of models, and that the two case studies produce genuinely different orderings among the top group rather than a fixed set regardless of clinical priority -- continues to hold under this second pass, though the specific models responsible for each pattern change. What does not survive is the first pass's specific headline finding that TabPFN~v2 wins by being strong-but-not-best on every super-metric: TabPFN~v2 is no longer a top-eight model in either case study, and the new co-leaders, TabPFN~2.5 and TabICL, do not distinguish themselves from one another on any super-metric at all, since their sub-metric profiles are now identical.


\section{Conclusions and Future Work}
\label{sec:conclusion}

The question this paper set out to answer is a practical one: given the growing number of foundation models for tabular data, how should a clinician, data scientist, or clinical informatics team decide which one to use? Existing surveys have cataloged what these models can do, but they do not provide a structured way to compare them against the specific demands of a real deployment context. \system{} is designed to fill that gap through a three-layer evaluation architecture.

At the base of that architecture, sub-metrics are operationally defined, measurable properties, each corresponding to a specific empirical test. Because running those tests systematically across all available models and datasets is itself a major research undertaking, this paper derives sub-metric scores from documented model design properties, a property-based approach that is transparent, auditable, and analogous to the use of structured expert consensus in clinical guideline development when direct trial evidence is not yet available. One layer up, six metrics aggregate sub-metrics into a six-dimensional profile covering generalizability, privacy, data efficiency, scalability, interpretability, and fairness, with domain-adjustable weights so that the same framework produces different, and more useful, rankings for a privacy-sensitive hospital deployment than for a research study using anonymized public data.

The framework also provides four supplementary compound scores (Transfer Learning Capacity, Model Resilience, Efficiency-Privacy Trade-off, and Clinical Trustworthiness) that aggregate sub-metrics across metric domains. These super-metrics do not replace the overall ranking as the basis for model selection; they serve as diagnostic complements that reveal how a top-ranked model performs across specific cross-domain concerns, such as readiness for multi-site deployment or clinical governance approval.

The two healthcare case studies demonstrate that the framework is not merely a theoretical construct. Working through the iron deficiency and heart failure use cases shows how domain-specific knowledge translates into concrete sub-metric weights, and how those weights produce model rankings that reflect the clinical context rather than generic benchmark performance. The same framework, applied with different weights, produces different rank orderings for models with genuine architectural trade-offs, making the basis for the recommendation fully transparent and traceable.

\subsection*{Limitations}

The sub-metric scores in this study are derived from documented model design properties using a property-based scoring approach, rather than from systematic empirical measurement. This is a deliberate choice: the property-based approach is transparent, accessible, and directly traceable, so a clinician or data scientist reviewing a model's score can identify which design properties drove it without needing access to a benchmarking pipeline. Precise measurement protocols for each sub-metric can be defined analytically and are retained in the framework's appendix definitions as commented specifications; future empirical work would replace the current property-based scores with directly measured values. The framework currently covers six evaluation dimensions; other considerations, such as energy consumption, robustness to distribution shift over time, or compliance with evolving AI regulation, may warrant inclusion as the field and its regulatory environment develop. The healthcare case studies focus on two specific clinical tasks, and extending the framework's application to other medical domains, and to domains outside healthcare entirely, remains to be demonstrated.

\subsection*{Future Work}

The most direct extension of this work is a large-scale empirical benchmarking study that uses \system{} as its evaluation protocol, collecting directly measured sub-metric scores for a wide range of models across diverse clinical datasets. Each sub-metric in the framework is defined with a specific measurement protocol in mind: IDT is measured as the average AUC drop across a held-out set of datasets from different institutions, MIR is measured by the success rate of a published membership inference attack, and DNR is measured by performance degradation as controlled noise is introduced. Replacing the property-based scores used in this paper with these directly measured values would allow systematic analysis of how sensitive the final rankings are to variations in sub-metric weights, and would establish whether the property-based approximations used here are conservative or optimistic relative to empirical ground truth.

The weighting scheme itself could be improved by replacing expert-assigned weights with a more systematic elicitation process. Multi-criteria decision-making methods such as the analytic hierarchy process (AHP) or preference learning from structured expert input would make the weight assignment more reproducible across different teams and institutions, and would provide a clearer audit trail for the ranking decisions.

Finally, as new tabular foundation models continue to emerge, the taxonomy presented in this paper will need to be maintained and updated. Integrating the framework with a living leaderboard or an automated model evaluation pipeline would allow practitioners to access up-to-date comparisons without waiting for each new model to be reviewed manually in a published study.

\section*{Acknowledgements}
M. Lotfian Delouee acknowledges support from the LabGPT project, funded by Amsterdam UMC Innovation Funding.

\printbibliography 

\appendix

\section{Appendix: Metric Definitions}
\label{sec:appendix}

The six metrics described in Section~\ref{sec:architecture} are each operationalized through a set of sub-metrics. This appendix provides the clinical and operational definition of each sub-metric: what question it is designed to answer, and what a higher or lower score means in practice. In this paper, sub-metric scores are derived from documented model design properties using the property-based scoring scheme described in Section~\ref{sec:healthcare}; the appendix entries describe what each sub-metric captures so that any score can be traced back to its clinical rationale. Figures~\ref{fig:tfm_heatmap_a} and~\ref{fig:tfm_heatmap_b} at the end of this appendix present the sub-metric scores for all 45 models surveyed in this work as color-coded heatmaps, making within-metric comparisons across the full model set immediately apparent.

\subsection{Generalizability}

Generalizability is measured through five sub-metrics, each probing a different dimension of how a model adapts when the data it encounters differs from what it was trained on.

\subsubsection*{Intra-Domain Transferability (IDT)}

IDT measures whether a model applies to a new dataset from the same clinical domain (e.g., a blood test dataset from a different hospital network) without retraining, by comparing performance across same-domain datasets against a reference benchmark. Higher IDT means more consistent cross-site performance, the property that lets a model trained at one hospital deploy at another without per-site retraining.

\subsubsection*{Cross-Domain Transferability (CDT)}

CDT extends this to transfer across domains, for instance from a clinical dataset to an administrative claims dataset, a harder test of whether a model has genuinely general representations rather than narrow, source-domain-specific ones. Higher CDT indicates broader cross-domain transfer capability and greater deployment flexibility.

\subsubsection*{Schema Robustness (SR)}

SR measures what fraction of structural variations between datasets (column ordering, semantically equivalent feature names such as ``Sex'' vs.\ ``Gender'', feature splits or merges, differing unit scales) a model handles without failing or degrading significantly. SR is the sub-metric most directly relevant to multi-site EHR deployments.

\subsubsection*{Feature Adaptation Capability (FAC)}

FAC measures how well a model maintains predictive accuracy when a training-time feature is missing at deployment, or when a new, previously unseen feature appears. High FAC means graceful rather than catastrophic degradation as the feature space changes, which matters where laboratory panels vary by clinical site.

\subsubsection*{Zero-Shot Inference Adaptability (ZIA)}

ZIA measures the fraction of completely new datasets on which a model achieves acceptable performance with no fine-tuning at all, relying entirely on representations learned during pre-training. High ZIA matters when a target site has no labeled data or rapid deployment is needed; it is the upper bound on how much transfer a model's pre-training has enabled.

\subsection{Privacy Preservation}

Privacy Preservation measures how well a model holds up against the range of attacks and regulatory compliance requirements that arise in clinical practice.

\subsubsection*{Data Anonymization Robustness (DAR)}

DAR measures how much performance a model retains when trained on anonymized rather than original data, since clinical data shared across institutions is routinely anonymized before distribution. Higher DAR indicates representations that do not rely on identifying signals anonymization removes, a precondition for practical multi-site clinical research.

\subsubsection*{Membership Inference Resistance (MIR)}

MIR quantifies how effectively a model resists membership inference attacks, where an adversary tries to determine whether a specific patient's record was part of the training set. Higher MIR means an attacker performs close to chance, so model outputs carry minimal information about any individual's inclusion.

\subsubsection*{Differential Privacy Compliance (DPC)}

DPC captures the balance between privacy strength (governed by the privacy parameter $\epsilon$, where smaller values mean stronger guarantees but more noise) and maintained predictive performance. A model with strong accuracy at a small $\epsilon$ scores high on DPC, which is particularly relevant under GDPR or HIPAA obligations.

\subsubsection*{Data Reconstruction Attack Resistance (DRAR)}

DRAR measures how difficult it is for an adversary to recover the actual contents of training records from a model's outputs or gradients, going further than membership inference. This is especially relevant for generative models, whose synthetic outputs may leak information about the records used to train them.

\subsubsection*{Third-Party Data Handling Compliance (TPDHC)}

TPDHC measures what fraction of relevant data-handling practices (encrypted transfer, anonymization before external services, consent management, GDPR/HIPAA adherence) a model's deployment architecture satisfies. A model that keeps computation on-premises and processes only de-identified data scores high; one that sends raw records to an external API without safeguards scores low and is unlikely to pass an information governance review.

\subsection{Data-Efficient Learning}

Data-Efficient Learning measures how much a model can achieve when the available training data is limited, sparse, or imperfect. This is a persistent challenge in clinical AI, where annotated datasets are often far smaller than those used for benchmarking.

\subsubsection*{Performance with Reduced Data (PRD)}

PRD measures how well a model maintains its accuracy as the training set shrinks. Higher PRD means the model retains most of its performance on a small fraction of the data, often the most practically relevant sub-metric for single-center deployments where labeled data is inherently limited.

\subsubsection*{Few-Shot Learning Capability (FLC)}

FLC measures the extreme case: performance with only a handful of labeled examples per class, directly relevant to rare clinical conditions, newly identified patient subgroups, and any setting where annotation cost makes large labeled datasets unrealistic. Higher FLC means effective generalization from very few examples even under high within-class variability.

\subsubsection*{Data Noise Robustness (DNR)}

DNR measures how well a model maintains its accuracy across the noise levels and types (missing labs, inconsistent coding, entry errors) common in routine clinical data but absent from cleaned benchmark datasets. Higher DNR means near-clean-data performance despite noisy real-world input, which is critical for reliable deployment in routine clinical environments.

\subsubsection*{Data Augmentation Sensitivity (DAS)}

DAS measures whether synthetic data augmentation genuinely helps a model, is neutral, or is harmful, when real labeled data is scarce. Higher DAS indicates genuine benefit from additional synthetic examples, which is relevant in settings such as federated learning where data collection is slow but synthetic generation is feasible.

\subsection{Scalability}

Scalability captures how well a model transitions across the wide range of dataset sizes and feature-space dimensionalities encountered in clinical practice, from single-department retrospective cohorts to national registries.

\subsubsection*{Resource Utilization Efficiency (RUE)}

RUE captures whether a model uses memory and compute proportionally as the dataset grows, or becomes increasingly expensive relative to the performance it returns. Higher RUE means more value extracted per unit of resource at scale; lower RUE signals the model may become impractical to operate at population scale.

\subsubsection*{Computational Scalability (CS)}

CS measures how a model's computational demand grows as dataset size increases. A model requiring disproportionately more compute for a linear increase in data is not practically scalable even with high absolute performance. Higher CS means compute grows at a manageable rate, determining whether a hospital's computational budget will scale reasonably as a screening model expands from one department to a population program.

\subsubsection*{Feature Scalability (FS)}

FS measures how well a model maintains predictive performance as the number of features increases, for instance as an EHR extract adds more laboratory values, vitals, or diagnostic codes. High FS means stability as the clinical team adds analytes or data sources; low FS means degradation or instability as feature dimensionality grows.

\subsubsection*{Training Time Efficiency (TTE)}

TTE captures whether additional training time actually improves performance, or whether a model plateaus quickly. Higher TTE means continued meaningful improvement as more data is processed, which determines whether periodic retraining as patient data accumulates is worth the computational investment.

\subsection{Interpretability}

Interpretability measures how well a model supports clinical human oversight by communicating not just what it predicted, but why, in terms a clinician can evaluate against their own knowledge and experience.

\subsubsection*{Feature Importance Clarity (FIC)}

FIC captures how well a model's feature attributions align with established clinical reference expectations. Identifying hemoglobin and ferritin as key drivers of an iron-deficiency prediction is credible, while an implausible administrative code invites justifiable skepticism. Higher FIC indicates stronger alignment between what the model highlights and what clinical experts would expect.

\subsubsection*{Model Transparency Score (MTS)}

MTS reflects a model's structural complexity adjusted for any built-in interpretability mechanisms, since comprehensible internal structure is easier to audit, explain to patients and ethics boards, and maintain over time. Models that are architecturally complex without compensating tools (e.g., sequential feature selection, built-in saliency maps) score low.

\subsubsection*{Explanation Consistency (EC)}

EC measures how stable and reproducible a model's explanations are across similar inputs. A model that explains near-identical patients very differently undermines clinical confidence even if each explanation looks plausible in isolation. Higher EC means more consistent explanations, essential for the explanations to function as reliable clinical decision support.

\subsubsection*{Local Interpretability (LI)}

LI evaluates the quality of patient-level explanations generated by post-hoc methods such as SHAP or LIME, measuring how faithfully they reflect a model's actual decision process for an individual prediction rather than merely approximating it. High LI is what most directly affects the trustworthiness of clinical decision support, where the explanation itself is part of what supports clinical judgment.

\subsection{Fairness and Bias Mitigation}

Fairness and Bias Mitigation measures how effectively a model avoids reproducing or amplifying the demographic disparities encoded in historical clinical training data.

\subsubsection*{Predictive Parity Bias Score (PPBS)}

PPBS quantifies the consistency of precision (positive predictive value) across demographic groups, since a high-risk prediction should be equally reliable regardless of the patient's group. Lower PPBS means some groups receive less trustworthy positive predictions, which in a clinical context means unnecessary follow-up or false reassurance.

\subsubsection*{Bias Amplification Score (BAS)}

BAS measures whether a model reduces, preserves, or amplifies the biases present in its training data, beyond simply reproducing them. Higher BAS means the model does not worsen existing disparities; models with explicit bias-mitigation mechanisms, such as adversarial debiasing or fairness-constrained training, tend to score higher.

\subsubsection*{Intersectional Fairness Score (IFS)}

IFS evaluates how equitably a model performs across intersectional subgroups (for example, age combined with sex and ethnicity) rather than single demographic characteristics, since the worst-off group is often defined by an intersection that single-group analyses miss. Higher IFS indicates consistent performance and error rates across all intersectional subgroups, which is the fairness dimension most relevant to equitable deployment in conditions such as iron deficiency and heart failure.

\subsection{Extended List of Tabular Foundation Models}

Figures~\ref{fig:tfm_heatmap_a} and~\ref{fig:tfm_heatmap_b} show the sub-metric profiles of all 45 tabular foundation models surveyed in this work. Each cell reports the score (range 4--10 under the audited property assignments; see Section~\ref{sec:audit-note} for why this range is narrower than an unaudited pass could produce) on one of the 25 sub-metrics defined in Sections~\ref{sec:appendix} above, grouped into the same six metric domains used throughout the paper: G\,=\,Generalizability (IDT, CDT, SR, FAC, ZIA), P\,=\,Privacy (DAR, MIR, DPC, DRAR, TPDHC), D\,=\,Data Efficiency (PRD, FLC, DNR, DAS), S\,=\,Scalability (RUE, CS, FS, TTE), I\,=\,Interpretability (FIC, MTS, EC, LI), and F\,=\,Fairness (PPBS, BAS, IFS). Metric band colors match those used in the healthcare heatmaps in Section~\ref{sec:healthcare}: dark green indicates a score of 10 (no property deficits), shading through yellow to red as deficits accumulate. Models are grouped by architectural paradigm, with prediction-focused and transfer/pretraining models in Panel~A, and generative, LLM-based, and NLP/QA-oriented models in Panel~B, and a heavier rule separating each group. Reading down any column gives an immediate ranking of all models on that sub-metric; reading across a row reveals a model's overall property profile.

\input{sections/fig_tfm_heatmap}

\end{document}